\documentclass[11pt,letterpaper]{article}

\usepackage{fullpage}
\usepackage{mathpazo}
\usepackage[numbers, compress, sort]{natbib}
\usepackage[utf8]{inputenc} % allow utf-8 input
\usepackage[T1]{fontenc}    % use 8-bit T1 fonts
\usepackage{hyperref}       % hyperlinks
\usepackage{url}            % simple URL typesetting
\usepackage{booktabs}       % professional-quality tables
\usepackage{amsfonts}       % blackboard math symbols
\usepackage{nicefrac}       % compact symbols for 1/2, etc.zhu
\usepackage{microtype}      % microtypography
\usepackage{caption}
\usepackage{amssymb}
\usepackage{graphicx}
\usepackage{threeparttable}
\usepackage{color}
\usepackage{complexity}
\usepackage{subcaption}
\usepackage{amsmath}
\usepackage{amsthm}
\usepackage{listings}
\usepackage{pifont}
\usepackage{tabularx}
\usepackage{fancyvrb}
\usepackage{comment}
\usepackage{multicol}
\usepackage{algorithm}
\usepackage{algorithmic}
\usepackage{setspace}
\usepackage{wrapfig}

\usepackage{float}
\usepackage{wrapfig}
\usepackage{mathtools}
\usepackage{thmtools,thm-restate}
\usepackage{paralist}
\usepackage{multirow}
\usepackage{scrextend}
\usepackage{enumitem}
\usepackage[noorphans]{quoting}
\usepackage{bm}
\usepackage{bbm}
\usepackage{thm-restate}
\usepackage{tabularx}
\usepackage{xcolor}
\usepackage{xspace}
\usepackage{times}
\usepackage[capitalize]{cleveref}
\theoremstyle{definition}

\renewcommand{\algorithmiccomment}[1]{\bgroup\hfill//~#1\egroup}

\definecolor{dkgreen}{rgb}{0,0.6,0}
\definecolor{gray}{rgb}{0.5,0.5,0.5}
\definecolor{mauve}{rgb}{0.58,0,0.82}

\title{\Large \textbf{On Active Privacy Auditing in Supervised Fine-tuning for White-Box Language Models}}
\date{}
\author{Qian Sun$^{1,3}$, Hanpeng Wu$^{1,3}$, Xi Sheryl Zhang$^{2,3 \thanks{Corresponding Author.}}$\\
	$^1$\ University of Chinese Academy of Sciences, Nanjing
	\\
        $^2$\ Institute of Automation, Chinese Academy of Sciences \\
        $^3$\ Nanjing Artificial Intelligence Research of AI \\
	\texttt{\{sunqian221, wuhanpeng23\}@mails.ucas.ac.cn},\\ \texttt{sheryl.zhangxi@gmail.com}
}

\begin{document}
	
	\maketitle
	
	\begin{abstract}
        The pretraining and fine-tuning approach has become the leading technique for various NLP applications. However, recent studies reveal that fine-tuning data, due to their sensitive nature, domain-specific characteristics, and identifiability, pose significant privacy concerns. To help develop more privacy-resilient fine-tuning models, we introduce a novel active privacy auditing framework, dubbed \textsc{Parsing}, designed to identify and quantify privacy leakage risks during the supervised fine-tuning (SFT) of language models (LMs). The framework leverages improved white-box membership inference attacks (MIAs) as the core technology, utilizing novel learning objectives and a two-stage pipeline to monitor the privacy of the LMs' fine-tuning process, maximizing the exposure of privacy risks. Additionally, we have improved the effectiveness of MIAs on large LMs including GPT-2, Llama2, and certain variants of them. Our research aims to provide the SFT community of LMs with a reliable, ready-to-use privacy auditing tool, and to offer valuable insights into safeguarding privacy during the fine-tuning process. Experimental results confirm the framework's efficiency across various models and tasks, emphasizing notable privacy concerns in the fine-tuning process. Project code available for \url{https://anonymous.4open.science/r/PARSING-4817/}
	\end{abstract}
	
	%%%%%%%%% BODY TEXT
	\section{Introduction}
	\label{sec:intro}
	
        Concerns regarding the privacy of training data pose a significant challenge in AI security, especially in sensitive domains like healthcare \cite{luo2022biogpt} and finance \cite{wu2023bloomberggpt}, where privacy issues are particularly pronounced. The study of privacy attacks and defense mechanisms for large language models (LLMs) is still in its early stages. In the past, research has primarily concentrated on extracting training data from pre-trained language models (PLMs) \cite{carlini2021extracting, lehman2021does, kandpal2022deduplicating}, leading to the development of novel theories regarding model memorization in LLMs \cite{carlini2022quantifying} and mechanisms of data leakage \cite{lee2023language}.

        Compared to the extensive and diverse datasets used during the pre-training stage, the training data in the fine-tuning phase presents much greater privacy risks. On one side, the constrained datasets employed for fine-tuning usually do not match the number of model parameters, resulting in an over-reliance on a limited set of data samples \cite{ding2023parameter, chen2023prompting, xiao2023large}. This reliance not only greatly impacts the model's ability to generalize, but also raises the likelihood of disclosing sensitive details from the training data. On the other side, fine-tuning datasets are frequently derived from particular real-world sectors and usually contain more intricate information. While improving performance on targeted tasks, this specificity also significantly increases the cost of privacy breaches.

	\begin{figure}[t]
          \centering
          \includegraphics[width=\linewidth]{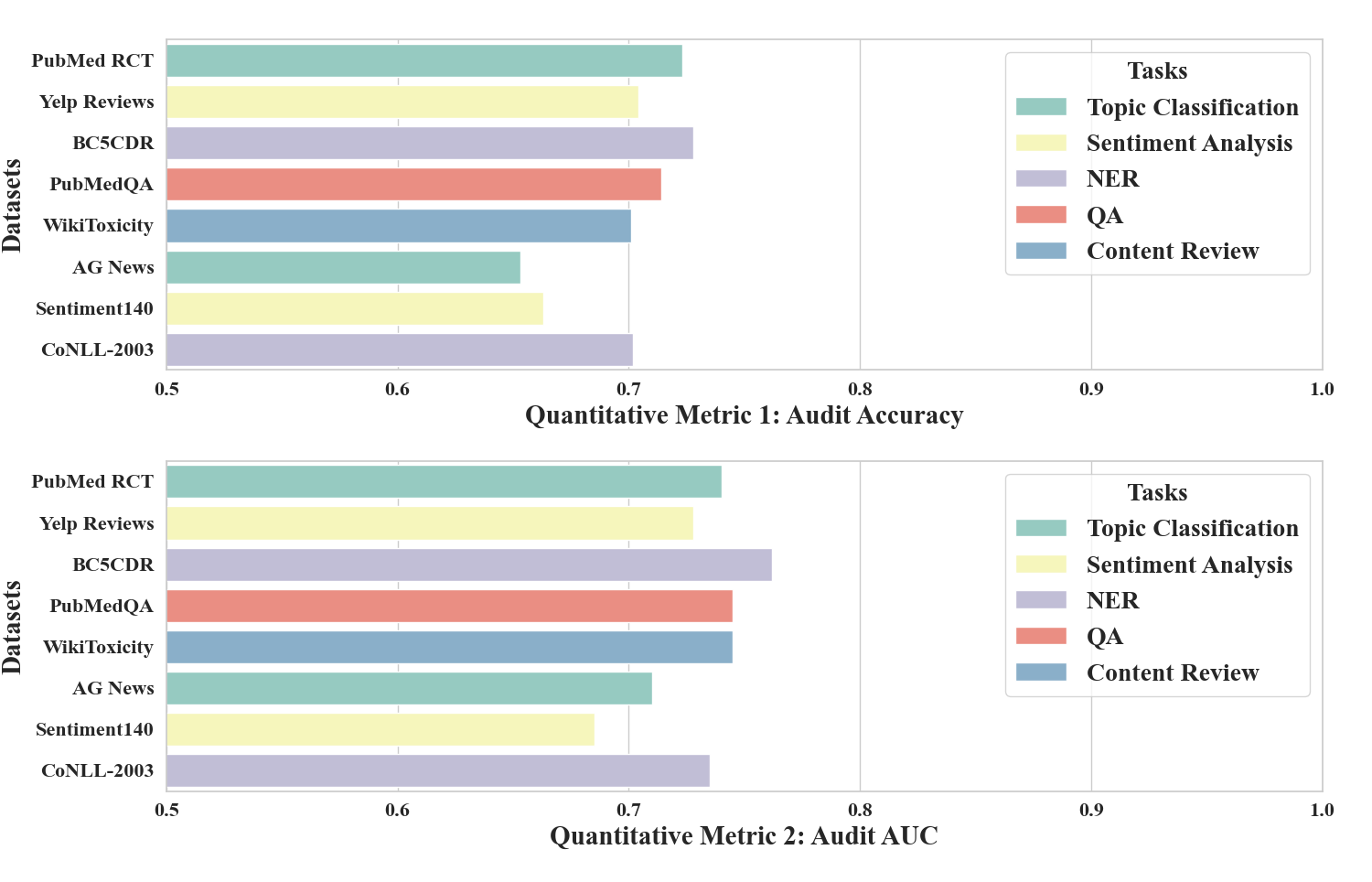}
          \caption{An example of privacy auditing results for fine-tuning a model on different tasks. Quantify the level of privacy leakage using two carefully designed metrics.}
        \end{figure}

        Thorough privacy evaluations greatly enhance user trust in models and their creators, serving as a crucial element in the sustainable advancement of technology. The emergence of advanced privacy attack algorithms \cite{chen2022amplifying, lee2023language} have increased the demand for innovative research on privacy protection and auditing. We believe that an effective privacy audit should be able to present the quantified results of privacy risks in a simple and intuitive manner, as shown in Figure 1. MIAs \cite{shokri2017membership, hu2022membership, mattern2023membership, gao2023similarity}, which form the basis for many other types of attack and privacy audit frameworks \cite{murakonda2020ml, carlini2022membership, nasr2018comprehensive, song2019auditing}, are employed as a critical benchmark to assess the privacy boundaries of models. These attacks provide a direct measure for assessing privacy risks, where the attack metrics may mirror the model's privacy level.

        Notably, the dynamic and intricate nature of the fine-tuning process arises both theoretical and practical challenges for privacy analysis under MIAs, hindering their successful application in deployed LMs. Additionally, existing privacy auditing methods primarily focus on the passive detection phase, where they assess privacy leakage after model training is completed. However, these methods often struggle to effectively address the dynamic changes occurring during the fine-tuning process, potentially overlooking critical privacy risks. To shed new light on developing more privacy-robust LMs, the following natural questions are explored as, \textit{Can the model fine-tuner actively identify privacy risks during the fine-tuning process? Can we quantify the privacy risks in this process using MIAs? What are the characteristics of privacy leakage during the process of LMs?} 

        As a result, it is necessary to develop more thorough methods to fill this technological gap. We note that non-members and members show noticeable variations in numerical data such as loss and gradient norms at the final layer, as well as in implicit data such as intermediate module outputs and gradients during fine-tuning. We have developed an effective methodological framework \textsc{Parsing} (Privacy Auditing on Risk of Supervised fine-tunING) to tackle these distinctions, highlighting and quantifying the privacy risks associated with the fine-tuning of LMs.

        \textbf{Contributions.} In this paper, we make the following contributions to the study of privacy in LMs.
        \begin{itemize}
        \item We introduce a novel privacy auditing framework applied during the fine-tuning phase of LMs, which has the potential to be employed in the foundation model designs. The \textsc{Parsing} framework aims to identify and quantify the risks of privacy leakage inherent in the fine-tuning process and seeks to help achieve a balance between privacy and utility. We offer a detailed explanation of the structure and operational mechanisms of the framework.
        
        \item We propose an \textit{active} two-stage white-box MIA method targeting LMs, applicable to various models including the GPT series and Llama series. By introducing a new methodological architecture and learning objectives, the method first optimizes the membership representation of samples, thereby enhancing their feature representation capability. This improvement significantly increases the effectiveness of MIAs on complex models.
        
        \item We evaluate \textsc{Parsing} on a range of models and diverse text tasks, and benchmarked it against existing studies. Empirical validation has shown that our framework is effective in detecting and quantifying privacy risks during the fine-tuning of LMs. Moreover, we also conduct a systematic analysis of the key factors leading to privacy vulnerabilities during the fine-tuning process, including task complexity, model size, text length, and so forth, in order to propose corresponding privacy protection strategies.
        \end{itemize}
	
	\section{Related Work}
	\subsection{Training Data Privacy Audits}

        Within a defined training environment and strategy for a particular model, the main objective of training data privacy audits \cite{nasr2021adversary, jagielski2020auditing, tramer2022debugging} is to identify and evaluate possible data privacy risks associated with the model, such as data leakage, unauthorized data inference, and the feasibility of adversaries reconstructing the original training data. It is widely accepted that a model service is considered to Privacy-Guaranteed if it does not divulge training data information during the inference phase. Particularly in domains sensitive to privacy, the development of models that guarantee privacy is of paramount importance. The privacy audit framework for training data \cite{murakonda2020ml, nasr2018comprehensive} is generally established on the basis of privacy attack and defense methods. It identifies and measures potential privacy risks using various attack techniques, and mitigates these risks through corresponding defense techniques.

        \textbf{Privacy Attack.} There are numerous privacy attack methodologies targeting model training data, notably including MIAs \cite{shokri2017membership, choquette2021label, nasr2018comprehensive}, training data extraction attacks \cite{carlini2021extracting, lee2023language, carlini2022quantifying}, model inversion attacks \cite{fredrikson2015model, inan2021training}, and property inference attacks \cite{ateniese2015hacking, ganju2018property}. (1) \textit{MIAs} are widely-studied privacy attack methods on basic neural network models. It aims to determine whether a specific sample was used in training a given model. While MIAs can pose significant privacy threats to models trained on sensitive data, they are also used as a tool to assess the privacy boundaries of a model \cite{tramer2022debugging, carlini2022membership, murakonda2020ml, nasr2021adversary}. (2) \textit{Training data extraction} attacks focus on extracting or reconstructing the original data used for training a model. Particularly in the realm of large language models, these attacks generally adhere to the generating-then-ranking methodology framework, as proposed by Chrlini et al. \cite{carlini2021extracting}. (3) The primary objective of \textit{model inversion attacks} is to infer aggregate details about the input data. These attacks are particularly effective when access to the model is limited, as they may reveal the model's internal structure or extract specific information about the training data. (4) \textit{Property Inference Attacks} aim at deducing sensitive attributes from the training data that are not directly revealed or are irrelevant to the task. For responsible data users and neural network model designers, mitigating Property Inference Attacks is an important consideration.

        \textbf{Privacy Defense.} To mitigate the risks introduced by these attack algorithms, researchers have developed various novel algorithmic techniques aimed at bolstering privacy safeguards. A commonly adopted approach is the incorporation of differential privacy \cite{dwork2014algorithmic} protocols during model training. This approach involves introducing noise at the data level \cite{cormode2018privacy} or algorithmic level \cite{abadi2016deep}, which helps shield personal information from precise identification. In addition to methods grounded in statistical theory with guaranteed robustness, practitioners also employ empirical and heuristic approaches, such as data anonymization \cite{murthy2019comparative}, data augmentation \cite{kaya2021does} and adversarial privacy enhancement \cite{jia2019memguard, nasr2018machine}. Although mathematical techniques offer robust privacy guarantees, they frequently result in reduced model performance and higher computational demands. Heuristic methods, which are typically simpler to apply, may lack the theoretical robustness provided by differential privacy. 

        \textbf{Privacy Measurement.} Privacy audit frameworks are pivotal in the development of secure artificial intelligence systems. Presently, methods for measuring privacy are primarily divided into two categories: static and dynamic methods. Static methods, based solely on privacy attacks \cite{murakonda2020ml, carlini2022membership, nasr2018comprehensive}, evaluate a model's vulnerability to privacy leakage post-training through simulated attacks. Typical techniques for these attacks include membership inference and training data extraction. Conversely, dynamic privacy measurement methods proactively monitor and assess privacy risks throughout the model's training process by integrating privacy measurement mechanisms into the training methodology \cite{carlini2019secret, jagielski2020auditing}. Compared to static methods, dynamic methods offer a more comprehensive analysis of privacy issues, thus providing researchers with enhanced insights for developing robust privacy protection strategies.

        \subsection{Membership Inference Attacks}
        The aim of membership inference attacks is to determine if a specific target sample was included in the training dataset of a designated model \cite{shokri2017membership, choquette2021label, nasr2018comprehensive, carlini2021extracting, hu2022membership}. Such attacks represent a notable risk to data privacy and have consequently attracted significant attention \cite{zhang2021membership, melis2019exploiting, yu2023bag, song2019auditing, hisamoto2020membership}.

        \textbf{Approach.} MIAs can be categorized based on the design approach of the attack model into two types: classifier-based techniques and metric-based techniques. Classifier-based methods aim to determine if a particular sample being analyzed is part of the training set of the target model by creating a binary classifier. Frequently employed methods include the use of shadow model training \cite{shokri2017membership}. Conversely, metric-based approaches distinguish between member and non-member instances by examining differences in various specific metrics. These metrics encompass prediction loss \cite{yeom2018privacy}, confidence vectors \cite{salem2018ml}, among others \cite{song2021systematic}.

        \textbf{Adversarial Knowledge.} Adversarial knowledge refers to the information an attacker has about a target model, which greatly impacts the effectiveness and complexity of the attack. In the majority of attacks, an attacker is typically restricted to the model's output data. However, in scenarios where the attacker is the model publisher, such as during privacy evaluations in the model training phase, the attacker may possess full knowledge of the model. Attacks are categorized based on the attacker's knowledge level into black-box and white-box attacks \cite{nasr2019comprehensive, hu2022membership}. In black-box attacks, the attacker can only access the model's confidence scores \cite{shokri2017membership} or just the hard labels \cite{choquette2021label, li2021membership, rahimian2020sampling}. Conversely, in white-box attacks, the attacker has access to detailed computational data regarding a particular sample \cite{nasr2018comprehensive}. This encompasses examining the model's internal architecture and operations, intermediate computation, gradient details, model loss, as well as all the information accessible in black-box situations. 

        \subsection{LLMs Fine-tuning} In recent years, the domain of NLP has experienced substantial advancements, the continuous enhancement of machine computational capabilities and data acquisition methods has heralded a new paradigm. This methodology is marked by comprehensive pre-training on general domain data, followed by precise fine-tuning for specific tasks \cite{zhao2023survey}. This development has been instrumental in the emergence of highly acclaimed and effective pre-trained language models, most notably the GPT series \cite{radford2018improving, radford2019language, ouyang2022training} and the Llama series \cite{touvron2023llama, zhao2023survey}.

        In the early stages of LMs fine-tuning, the predominant approach was full-parameter fine-tuning (FFT), i.e., all parameters of the pre-trained model were adjusted for downstream tasks. However, this method often requires substantial data to tune a large number of parameters, and as the model size increases, so does the demand for computational resources. Consequently, parameter-efficient fine-tuning techniques \cite{peft} have emerged. Among these, popular methods include prompt tuning \cite{lester2021power, liu2021p} and prefix tuning \cite{li2021prefix}, where only a subset of the model's parameters is fine-tuned for downstream tasks, or more targeted adjustments are made to specific model components. 
        
        Adapter-based methods \cite{houlsby2019parameter, lin2020exploring, pfeiffer2020adapterfusion} represent another approach, in which adapter structures are designed and embedded in the transformer architecture, and only the newly added adapter structures are fine-tuned during training. However, the former can pose challenges in optimization, while the latter may introduce inference latency \cite{hu2021lora}. In response to these challenges, researchers have proposed more advanced low-rank adaptation methods. These methods aim to confine changes in model weights to a low-rank subspace \cite{hu2021lora}. By decomposing the weight matrices of the model, only a small subset of parameters needs to be fine-tuned. This approach not only minimizes computational consumption, but also retains sufficient model flexibility to learn new tasks. Notable updated methods include AdaLoRA \cite{zhang2023adaptive}, LoftQ \cite{li2023loftq}, among others.
	
	\section{Preliminaries}
        At a colloquial level, MIAs provide a means of quantifying privacy leakage risks, forming the foundation for many other attacks and privacy audit frameworks. They are also often used as a benchmark for evaluating privacy protection measures. In our study, we focus more on employing the MIA method as the core component of the auditing framework to identify privacy risks during the model fine-tuning phase. The auditor is the entity executing the attack. In this section, we outline the basic settings of the MIA component.

        \subsection{Capability of the Auditor} 
        In this scenario, we examine an attack launched by the entity engaged in fine-tuning, meaning \textit{the auditor is actively participating in the fine-tuning procedure}. Consequently, the auditor can extensively utilize the model's internal details to deduce its training data. This capability enables an assessment of the highest possible privacy leakage from the fine-tuned dataset. 

        Regarding data knowledge, the auditor has extensive knowledge of the fine-tuning dataset $D_{ft}$, encompassing both instances and distributions. In terms of model knowledge, the auditor has complete access to the model $M_{ft}$ after each fine-tuning epoch, including its parameters $\boldsymbol{\theta}_{ft}$ and the model architecture. The auditor possesses an understanding of the loss functions $\mathcal{L}$ and optimization methods from the training algorithm standpoint. The auditor's knowledge is denoted as $\mathcal{K}$.

        \subsection{Objective of the Auditor} 
        \textbf{Definition 1.} \textit{(Classifier-Based Membership Inference Attack) Consider a target model $M_{ft}$ trained on a fine-tuned training dataset $D^{ft}_{train}$, and an attacker possessing knowledge $\mathcal{K}$. Given an instance sample $x$, the membership inference module $A$ is a function that maps the input tuple $(x,M_{ft},\mathcal{K})$ to a binary decision, defined as:
        \begin{equation}
          A(x,M_{ft},\mathcal{K}) \rightarrow \{0, 1\},
        \end{equation}
        where $1$ indicates that the sample $x$ is inferred to be a member of the training dataset $D^{ft}_{train}$, while $0$ indicates otherwise.}
        
        The auditor's objective is to ascertain whether sample $x$ is present in the fine-tuning training dataset $D^{ft}_{train}$. 

        \subsection{Quantitative Metrics} 
        To thoroughly assess \textsc{Parsing}'s effectiveness and achieve the goal of quantifying privacy, we utilize three quantitative metrics: balanced attack accuracy \cite{shokri2017membership, nasr2018comprehensive}, TPR at low FPR \cite{carlini2022membership} and AUC. And report the auditing results using the ROC curve presented on a logarithmic scale.
        
        In an effort to expose privacy leakage risks, the balance accuracy metric is sufficient. In this context, ``balanced" means that the training and testing datasets for audit models contain an equal number of member and non-member samples. However, for MIA assessment, the TPR at low FPR is a more appropriate metric. The metric aims to report the attack's true positive rate at a fixed low false positive rate. In MIAs, the emphasis is on instances identified as member samples, as opposed to non-member samples. Accuracy by itself cannot capture the relationship between the TPR and the FPR. The ROC curve offers a more detailed insight into this equilibrium. Utilizing a logarithmic scale can emphasize performance at low false positive rates.

        \begin{figure*}[t]
          \centering
          \includegraphics[width=\linewidth]{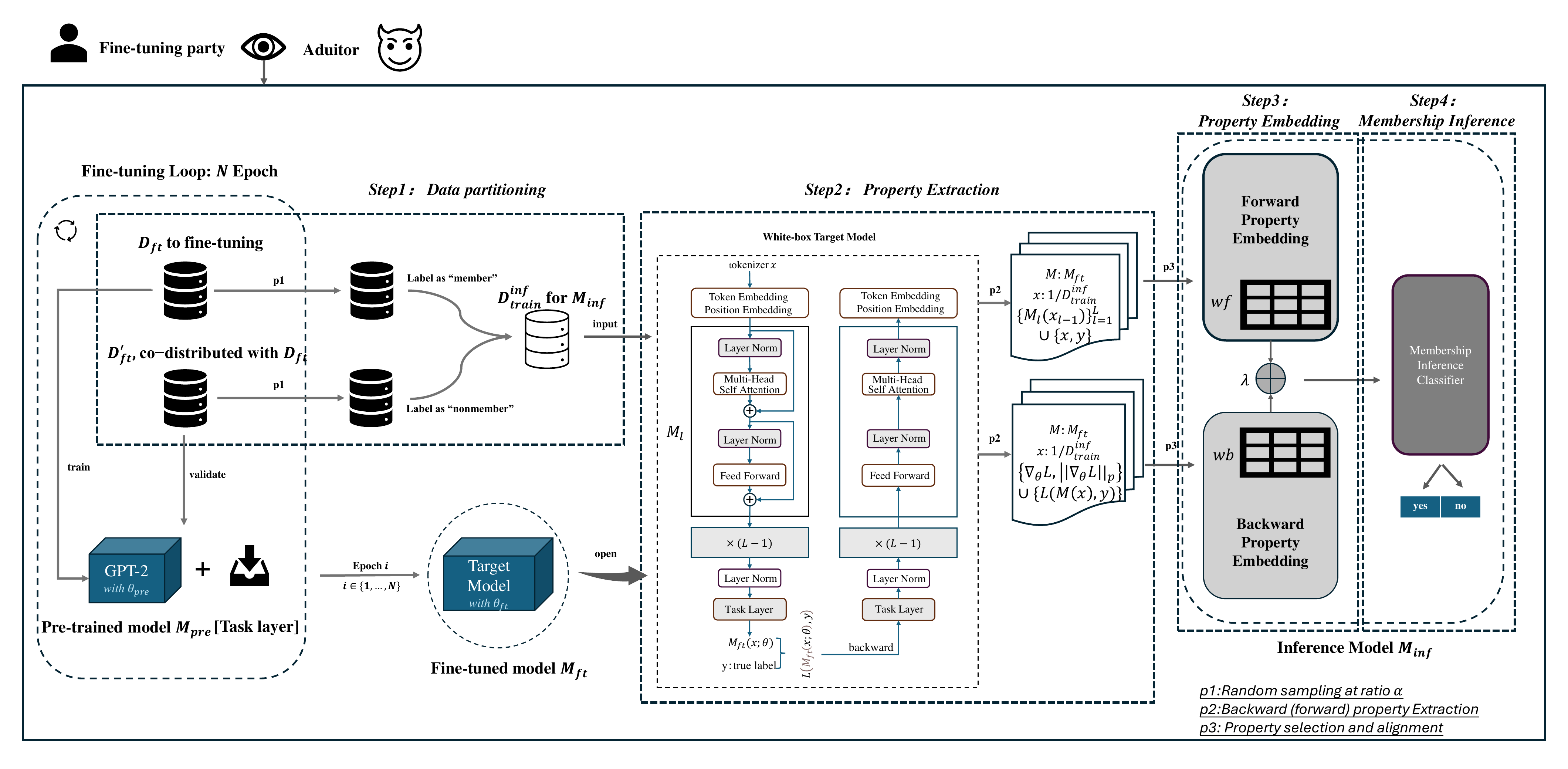}
          \caption{A comprehensive breakdown of the auditing framework \textsc{Parsing} embedded in the model fine-tuning process based on white-box MIAs, including data partitioning, property extraction, property embedding, and membership inference.}
        \end{figure*}
        
	\section{Design of \textsc{Parsing}}
	\label{sec:design}
	\noindent
        The complex and dynamic nature of fine-tuning LMs makes simple information extraction methods insufficient for capturing the detailed patterns learned from the training data. We concentrate on learning high-quality property embeddings from the white-box information of diverse samples, allowing the inference module to more effectively differentiate between non-member and member samples. In this section, we present a detailed overview of the principles and workflow of \textsc{Parsing}. Figure 2 provides a detailed breakdown of the components of the framework.
        
        \subsection{Data Partitioning}
        The auditor seeks to conduct a privacy examination on the target model during its fine-tuning stage. To achieve this, it is essential to divide a dataset $D_{inf}$ from $D_{ft}$ for audit before starting the fine-tuning and assign labels to these data. The original training dataset for model fine-tuning is defined as $D^{ft}_{train} =  \{(x_i, y_i) | i = 1, 2, ..., N\}$, where $N$ is the total number of samples in the dataset. Determine the scaling factor $\alpha$, and randomly partition two disjoint subsets from $D^{ft}_{train}$, and label them as $D^1_{inf} =  \{(x_j, y_j, 1) | j = 1, 2, ..., \alpha{N}\}$, where $1$ is the membership label, constituting the member samples in the training and test sets of the inference module. $D^0_{inf} = \{(x_j, y_j, 0) | j = 1, 2, ..., \alpha{N}\}$ is collected from data that is distributed identically to $D^{ft}_{train}$, comprising the non-member samples in the training and test sets of the inference model. Based on the above, the training set $D^{inf}_{train}$ and the test set $D^{inf}_{test}$ are constructed for the following modules.
        
        \subsection{Property Extraction}
        \textit{Property} refers to the intermediate computation results of the LMs on specific data, which we consider as properties of the data for the model. 
        During the fine-tuning phase, the model is dissected to extract intermediate computation, which is then compiled into a repository of candidate properties for optimization of property embeddings. We categorize the properties into two groups: forward properties and backward properties. In particular, the definition is given as,
        
        \textit{Forward property} includes the results from each intermediate module, such as the token embedding, attention, and task-specific blocks. Given an input sample $x$ and its label $y$, for each intermediate module $\mathcal{M}_{l}$, where $1 \leq l \leq L$ and $L$ denotes the total number of intermediate modules in the model, compute its output $x_{l+1}$ to be used as the input for the subsequent block, which can be expressed as:
        \begin{equation}
          x_{l+1} = \mathcal{M}_{l}(x_{l}).
        \end{equation}
        
        For the entire target model, all forward properties is formalized as:
        \begin{equation}
          \mathcal{I}_{f} = \left\{\mathcal{M}_{l}(x_{l-1})\right\}^L_{l=1} \cup \left\{x, y\right\}.
        \end{equation}
        
        \textit{Backward property} permits gradient computation that includes the calculated gradients $\nabla_{\boldsymbol{\theta}} \mathcal{L}\left( {M}_{ft}({x}),y\right)$ and their norms $\|\nabla_{\boldsymbol{\theta}} \mathcal{L}\|_{p}$ for each model parameter, calculated from the loss $\mathcal{L}\left({M}_{ft}({x}),y\right)$. Here, $\boldsymbol{\theta}$ denotes the set of all the parameters of the target model. The backward property repository is defined as:
        \begin{equation}
          \mathcal{I}_{b} = \left\{\nabla_{\boldsymbol{\theta}} \mathcal{L}, \|\nabla_{\boldsymbol{\theta}} \mathcal{L}\|_{p}, \mathcal{L}\left( {M}_{ft}({x}),y\right)\right\}.
        \end{equation}
        
        The white-box candidate property repository is formalized as follows:
        \begin{equation}
          \mathcal{I} = \mathcal{I}_{f} \cup \mathcal{I}_{b}.
        \end{equation}
        
        \subsection{Property Embedding}
        \textit{Property embedding} step transforms the properties extracted from the target model into feature representations used for privacy evaluation. The property embedding module is the central element of \textsc{Parsing}.\\
        \textbf{Definition 2.} \textit{(Embedding with Forward and Backward Properties) Let $\mathcal{I}$ be a candidate property set. Forward properties $\mathcal{I}_{f}$ and backward properties $\mathcal{I}_{b}$ are selected from $\mathcal{I}$, ensuring alignment through a normalization and dimension-matching process. Separate embedding generators are employed for forward and backward properties, and these are connected via learnable parameters. The representation embedding $r(x)$ for a sample $x$ is defined as:
        \begin{equation}
          r(x) = \lambda \mathbf{R}_{wf}\left[\mathcal{F}_{f} \left( \mathcal{I}_{f} \right)\right] \oplus (1-\lambda)\mathbf{R}_{wb}\left[ \mathcal{F}_{b} \left(\mathcal{I}_{b} \right) \right],
        \end{equation}
        where $\mathcal{F}$ denotes the properties alignment process, $\mathbf{R}$ represents the property embedding generator, $wf$ and $wb$ are the learnable parameters associated with the forward and backward embedding generators, $\oplus$ signifies the concatenation operation, and $\lambda$ is the parameter for information weighting, thus deriving the embedding $r(x)$ for sample $x$.}
        
        The learner generates member embeddings $r_{mem}$ or non-member embeddings $r_{non}$ for each sample point through one-dimensional convolution and fully connected layers. The goal is to increase the difference in embeddings between member and non-member samples, while maintaining or increasing the similarity within member samples. However, unlike the triplet loss or contrastive loss, MIA is not a symmetrical problem; we are more concerned with the internal similarity of member samples; see Appendix D for details.
        
        Let $\mathcal{S}_{\text{mem}}$ be the set of member sample embeddings and $\mathcal{S}_{\text{non}}$ be the set of non-member sample embeddings. We integrate the tasks as follows:
        \begin{equation}
           L = L_{d}(\mathcal{S}_{\text{mem}}, \mathcal{S}_{\text{non}}) +\mu(t){L}_{s}(\mathcal{S}_{\text{mem}}),
        \end{equation}
        herein, $L_{d}$ measures the representational difference between member and non-member samples. $L_{d}$ can be expressed as:
        \begin{equation}
          L_{d} = \frac{1}{|\mathcal{S}_{mem}|}\sum_{i=1}^{|\mathcal{S}_{mem}|} \max \left[0,m-d \left(r_{\text{mem},i},r_{\text{non},i} \right)\right],
        \end{equation}
        where $m$ is a boundary parameter, $|\mathcal{S}_{mem}|$ indicates the number of member embeddings. In our method, the distance $d$ is specifically defined as the cosine similarity between two vectors. $\mu(t)$ is a weight function, balancing the importance of these two metrics:
        \begin{equation}
          \mu(t)=\mu_{0} \cdot \exp(-k \cdot t),
        \end{equation}
        where $\mu_0$ denotes the initial weight, and $k$ is the decay coefficient. $L_{s}$ is the concentration loss of $\mathcal{S}_{\text{mem}}$, used to maintain the internal consistency of member samples:
        \begin{equation}
          L_{s} = \frac{1}{|\mathcal{S}_{mem}|}\sum_{i,j=1}^{|\mathcal{S}_{mem}|} d(r_{mem,i},r_{mem,j}).
        \end{equation}
        Its primary purpose is to ensure that the representations of member samples in the embedding space are as similar as possible, thereby improving the accuracy of member inference.

        In practice, we typically care more about samples predicted to be members, or rather, we aim to find the internal similarity of member samples. The internal representation of non-member samples is less of a concern, as long as their interclass distance from member samples is sufficiently large. In privacy audit tasks, we do not assign equal attention to the samples predicted as 0 and 1. For details and arguments, see Appendix.
        
        \subsection{Membership Inference}
        
        The inference module, acting as the classifier, is trained using sample property embeddings $r$ and explicit membership labels $y_{0,1}$. This component is structured as a multilayer fully connected neural network: the input layer processes the embeddings as features; the hidden layers explore the intricate relationships between these features; and the output layer determines membership status of the samples. For each sample's embedded representation $r(x)$, the classifier predicts its membership label $\hat{l}$, and the cross-entropy loss is:
        \begin{equation}
          L_{ce} = -\frac{1}{N}\sum_{i} [l_ilog(\hat{l}_i)+(1-l_i)log(1-\hat{l}_i)],
        \end{equation} 
        where $l$ is the true membership label, and $\hat{l}$ is the predicted label. Cross-entropy loss is used to train the classifier to accurately distinguish member samples from non-member samples. Combine the losses of the two components for joint optimization, The final optimization objective combines the aforementioned losses:
        \begin{equation}
          L_{total}=L+\nu L_{ce},
        \end{equation} 
        where $\nu$ represents the weight parameter, balance the influence of each part's loss.  
        
        The property embeddings generated in the previous steps have been fine-tuned to equip the classifier with essential information for differentiating between member and non-member samples, significantly enhancing the precision of membership inference. The classifier is trained by minimizing the cross-entropy loss between the predicted membership labels and the actual labels $y_{0,1}$.

        \section{Application and Experiments}
        \begin{table}
            \centering
            \setlength{\tabcolsep}{2.5mm}
            \fontsize{10pt}{14pt}\selectfont
            \begin{tabular}{l|ccc|ccc} 
                \toprule
                \multirow{2}{*}{Datasets}& \multicolumn{3}{c|}{Balanced accuracy} & \multicolumn{3}{c}{AUC} \\
                \cline{2-7} 
                & $A_{loss}$ & $A_{black}$ & $\textsc{Parsing}$ & $A_{loss}$ & $A_{black}$ & $\textsc{Parsing}$\\
                \midrule
                PubMed\_RCT & 0.719$\pm$0.005 & 0.668$\pm$0.003 & \textbf{0.741$\pm$0.012} & 0.745$\pm$0.008 & 0.751$\pm$0.007 & \textbf{0.775$\pm$0.008}\\
                Yelp Reviews & 0.704$\pm$0.005 & 0.643$\pm$0.010 & \textbf{0.723$\pm$0.012} & 0.722$\pm$0.010 & 0.736$\pm$0.013 & \textbf{0.755$\pm$0.015}\\
                BC5CDR  & 0.723$\pm$0.004 & 0.678$\pm$0.014 & \textbf{0.742$\pm$0.004} & 0.748$\pm$0.013 & 0.750$\pm$0.011 & \textbf{0.769$\pm$0.010}\\
                PubMedQA & 0.706$\pm$0.007 & 0.692$\pm$0.012 & \textbf{0.765$\pm$0.019} & 0.738$\pm$0.011 & 0.760$\pm$0.018 & \textbf{0.794$\pm$0.020}\\
                Wiki Toxicity & 0.702$\pm$0.003 & 0.664$\pm$0.003 & \textbf{0.717$\pm$0.013} & 0.737$\pm$0.010 & 0.742$\pm$0.013 & \textbf{0.758$\pm$0.012} \\
                \cline{1-7} 
                \rule{0pt}{10pt}AG News & 0.687$\pm$0.003 & 0.652$\pm$0.013 & \textbf{0.700$\pm$0.013} & 0.715$\pm$0.013 & 0.701$\pm$0.013 & \textbf{0.734$\pm$0.012}\\
                Sentiment140 & 0.662$\pm$0.004 & 0.619$\pm$0.005 & \textbf{0.683$\pm$0.003} & 0.685$\pm$0.005 & 0.656$\pm$0.012 & \textbf{0.703$\pm$0.015}\\
                CoNLL-2003 & 0.709$\pm$0.002 & 0.679$\pm$0.008 & \textbf{0.721$\pm$0.014} & 0.744$\pm$0.009 & 0.757$\pm$0.014 & \textbf{0.777$\pm$0.013}\\
                \bottomrule
            \end{tabular}
            \caption{The maximum values achieved by \textsc{Parsing} and baselines in terms of \textit{balanced accuracy} and \textit{auc} when fine-tuning the GPT-2 XL model on different text tasks.}
        \end{table}
        
        \begin{table}
            \centering
            \setlength{\tabcolsep}{5mm}
            \fontsize{10pt}{14pt}\selectfont
            \begin{tabular}{l|ccc} 
                \toprule
                \multirow{2}{*}{Datasets}& \multicolumn{3}{c}{TPR at 0.1 FPR} \\
                \cline{2-4} 
                & $A_{loss}$ & $A_{black}$ & $\textsc{Parsing}$ \\
                \midrule
                PubMed\_RCT & 0.258$\pm$0.042 & 0.333$\pm$0.075 & \textbf{0.371$\pm$0.080}\\
                Yelp Reviews & 0.235$\pm$0.057 & 0.298$\pm$0.085 & \textbf{0.333$\pm$0.016}\\
                BC5CDR  & 0.264$\pm$0.043 & 0.320$\pm$0.039 & \textbf{0.374$\pm$0.011}\\
                PubMedQA & 0.274$\pm$0.040 & 0.368$\pm$0.092 & \textbf{0.406$\pm$0.063}\\
                Wiki Toxicity & 0.235$\pm$0.014 & 0.301$\pm$0.051 & \textbf{0.339$\pm$0.077} \\
                \cline{1-4} 
                \rule{0pt}{10pt}AG News & 0.221$\pm$0.033 & 0.282$\pm$0.067  &  \textbf{0.302$\pm$0.053}\\
                Sentiment140 & 0.201$\pm$0.050 & 0.238$\pm$0.077 & \textbf{0.262$\pm$0.041}\\
                CoNLL-2003 & 0.253$\pm$0.923 & 0.314$\pm$0.067 & \textbf{0.341$\pm$0.073}\\
                \bottomrule
            \end{tabular}
            \caption{The maximum values achieved by \textsc{Parsing} and baselines in terms of \textit{TPR at 0.1 FPR} when fine-tuning the GPT-2 XL model on different text tasks.}
        \end{table}
        
        In this section, we empirically evaluate the effectiveness of $\textsc{Parsing}$ in privacy auditing during model tuning. Empirically demonstrated the privacy vulnerabilities of fine-tuning LMs and analyzed the nature of their privacy leakage.
        
        \subsection{Experimental Setup}
        \textbf{Targeted Models}
        We have selected GPT-2 \cite{radford2019language}, GPT-Neo \cite{black2021gpt}, and Llama2 \cite{touvron2023llama} as the focus of our study. The primary reason for the selection is their extensive use in various tasks, making them essential subjects for investigating vulnerabilities in privacy leakage. Moreover, these models vary in size, allowing us to explore how their architecture and capacity influence their susceptibility to attacks. For every text task, we enhance the model with an optional task-specific layer. 
        
        \textbf{Tasks and Datasets} We assessed our approach on a range of text-related tasks, including topic classification, NER and QA, among others. The datasets utilized comprise PubMed\_RCT, BC5CDR, PubMedQA, etc. Table 1 and Table 2 list all the experimental datasets.
        
        \textbf{Baselines} Since there is currently no privacy audit algorithm specifically designed for the fine-tuning process of language models, we focus on alternative MIA methods that can replace the components of our framework, without disrupting the model structure or imposing additional operations on the model. \cite{nasr2019comprehensive} proposed a white-box MIA method for visual models, also based on extracting information. We replicated this method, adhering to its best information utilization (i.e., based on the output of the last layer), and named it $A_{black}$. The loss-based MIA $A(x,\theta)=-\xi (x,\theta)$ proposed in \cite{yeom2018privacy} is still the most commonly used method, named as $A_{loss}$. To scientifically compare our method, we optimized algorithms and adjusted frameworks for the selected baselines. The improved methods outperformed the original algorithms on the data used in this study. 
        
        Detailed experimental setups, hyperparameters, data partitioning, etc., can be found in Appendix.
        
        \subsection{Results Analysis}
        
        \begin{figure}[t]
             \centering
             \begin{subfigure}[b]{0.47\textwidth}
                 \centering
                 \includegraphics[width=\textwidth]{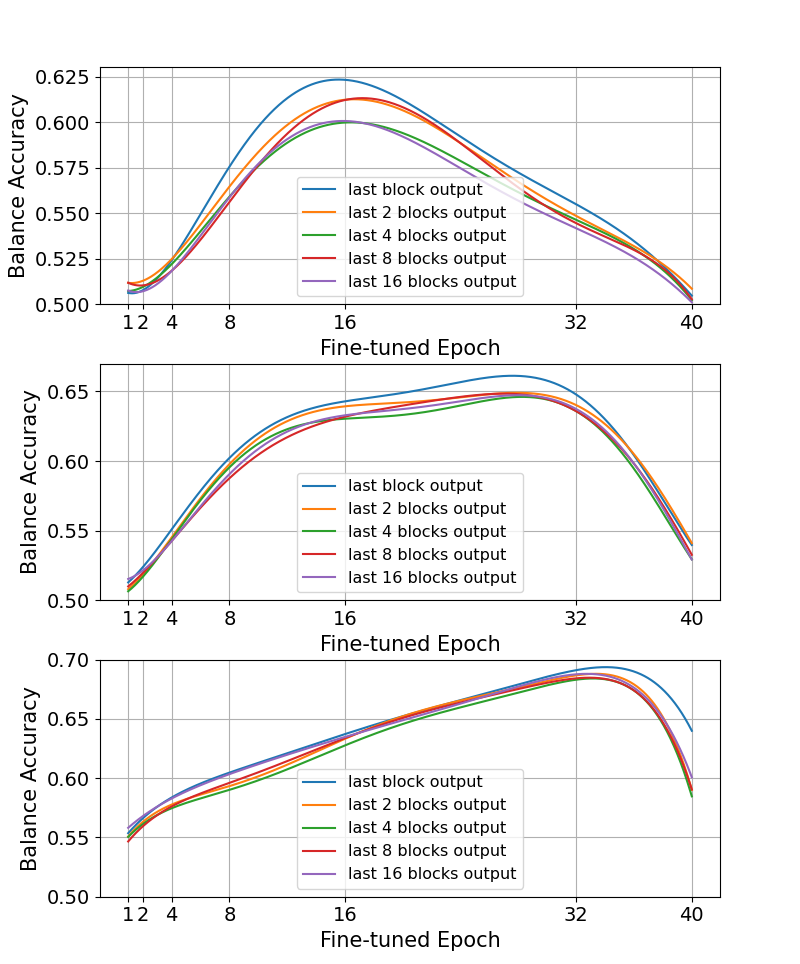}
                 \caption{forward properties}
                 \label{fig:forward information}
             \end{subfigure}
             \hspace{0.01\textwidth}
             \begin{subfigure}[b]{0.47\textwidth}
                 \centering
                 \includegraphics[width=\textwidth]{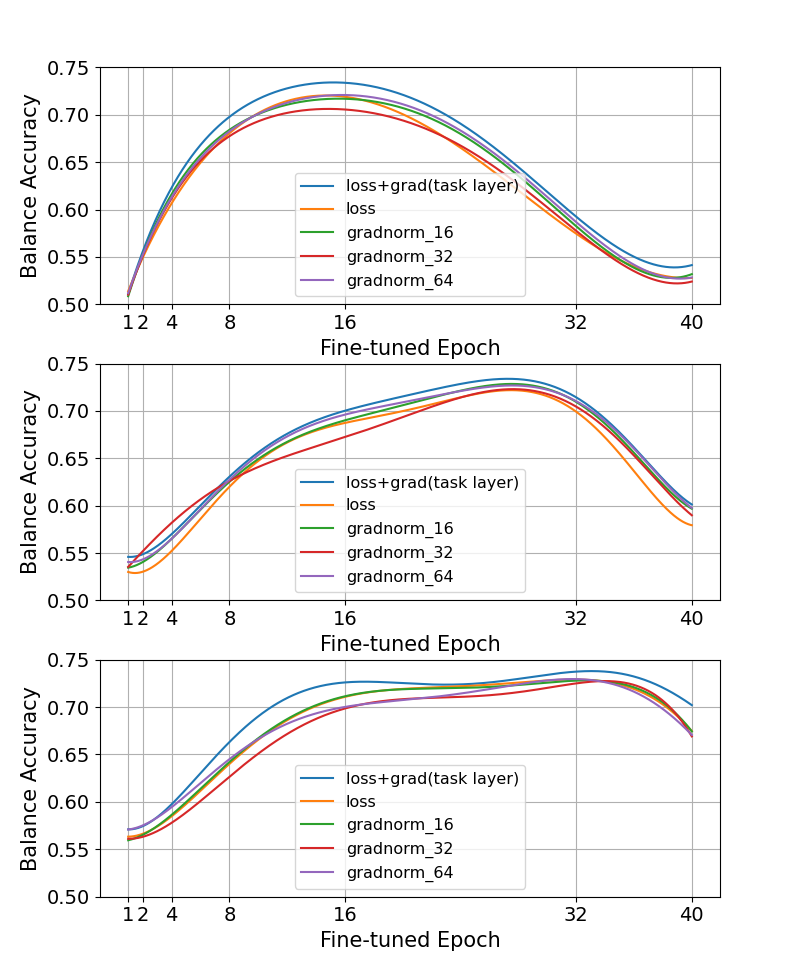}
                 \caption{backward properties}
                 \label{fig:backward information}
             \end{subfigure}
                \caption{Balance accuracy variation curve of the audit across three model sizes on PubMed-RCT dataset over 40 fine-tuning epochs (from top to bottom: GPT-2-medium, GPT-2-large, GPT-2-xl)}
                \label{fig:two graphs}
        \end{figure}
        
        \textbf{Privacy Risk Quantification.} Our aim is to reveal risks as thoroughly as possible and quantify the privacy risks. Tables 1 and 2 display the peak values achieved by \textsc{Parsing} on the metrics of balanced accuracy, AUC and TPR at 0.1 FPR (TPR$_{0.1}$) when fine-tuning the GPT-2 XL model for various tasks. The experiments performed on other models are detailed in Appendix G. Under random circumstances, the performance of each metric is 0.5, 0.5 and 0.1 respectively; higher values indicate greater privacy risks. 
        
        Using these metrics, we can measure and contrast the privacy risks involved in the model fine-tuning process. For example, the highest privacy risk associated with fine-tuning PubMedQA on the GPT-2 XL model can be measured as: $balance\_acc=.765$; $auc=.79$; TPR$_{0.1}=0.403$. The experimental results presented in the table indicate that fine-tuning LMs is particularly vulnerable to privacy breaches.

        \textbf{Text Tasks and Instance Length.} In addition to showing that fine-tuning LMs can result in notable privacy leakage risks, Tables 1 and 2 also indicate at least two conclusions when the fine-tuning data are scarce or restricted.
        \begin{itemize}
            \item The probability of privacy breaches is roughly connected to the task's complexity. Although QA tasks show lower privacy leakage risks at the beginning of fine-tuning, as training advances, the peak values of the measured metrics exceed those of simpler tasks such as topic classification; refer to Appendix. 
            \item The privacy risk is linked to the length of the training text. The top section of Table 1 displays long text tasks, whereas the bottom presents short text tasks. It is clear that, for similar text tasks, long text tasks are more prone to privacy violations following overtraining, despite the common belief that longer texts are simpler to generalize.
        \end{itemize}
        
        One possible reason is that, although simpler tasks are more susceptible to overfitting, the usual fine-tuning datasets remain insufficiently large compared to the model's vast number of parameters. Complex tasks require a more thorough understanding and examination of the data. For instance, for QA tasks, the model needs to not only comprehend and remember the content of paragraphs but also identify and retrieve relevant information within the context. Sometimes, it must also reason and integrate various pieces of information to generate answers, resulting in a deeper retention of the data. Compared to short texts, longer texts provide more extensive contextual information and may include additional features, which makes it simpler for the model to recall and extract properties.
        
        The analysis was performed on restricted datasets, and the findings could still be influenced by data distribution, the architecture of the target model, and other factors. However, current experiments cannot eliminate these impacts, indicating the need for additional research and dialogue.
        
        \begin{figure}[t]
             \centering
             \begin{subfigure}[b]{0.45\textwidth}
                 \centering
                 \includegraphics[width=\textwidth]{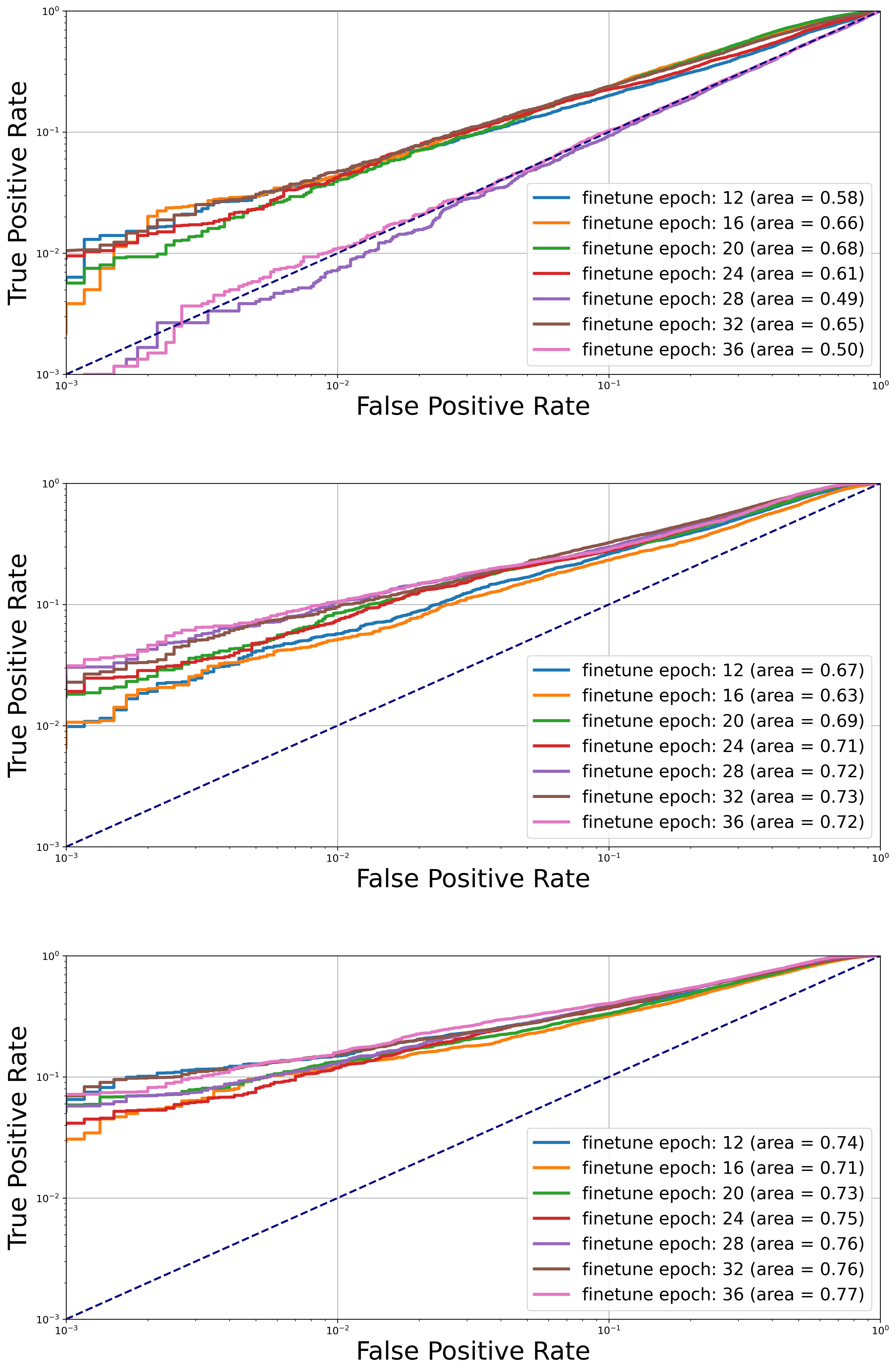}
                 \caption{forward properties}
                 \label{fig:forward information}
             \end{subfigure}
             \hspace{0.05\textwidth}
             \begin{subfigure}[b]{0.45\textwidth}
                 \centering
                 \includegraphics[width=\textwidth]{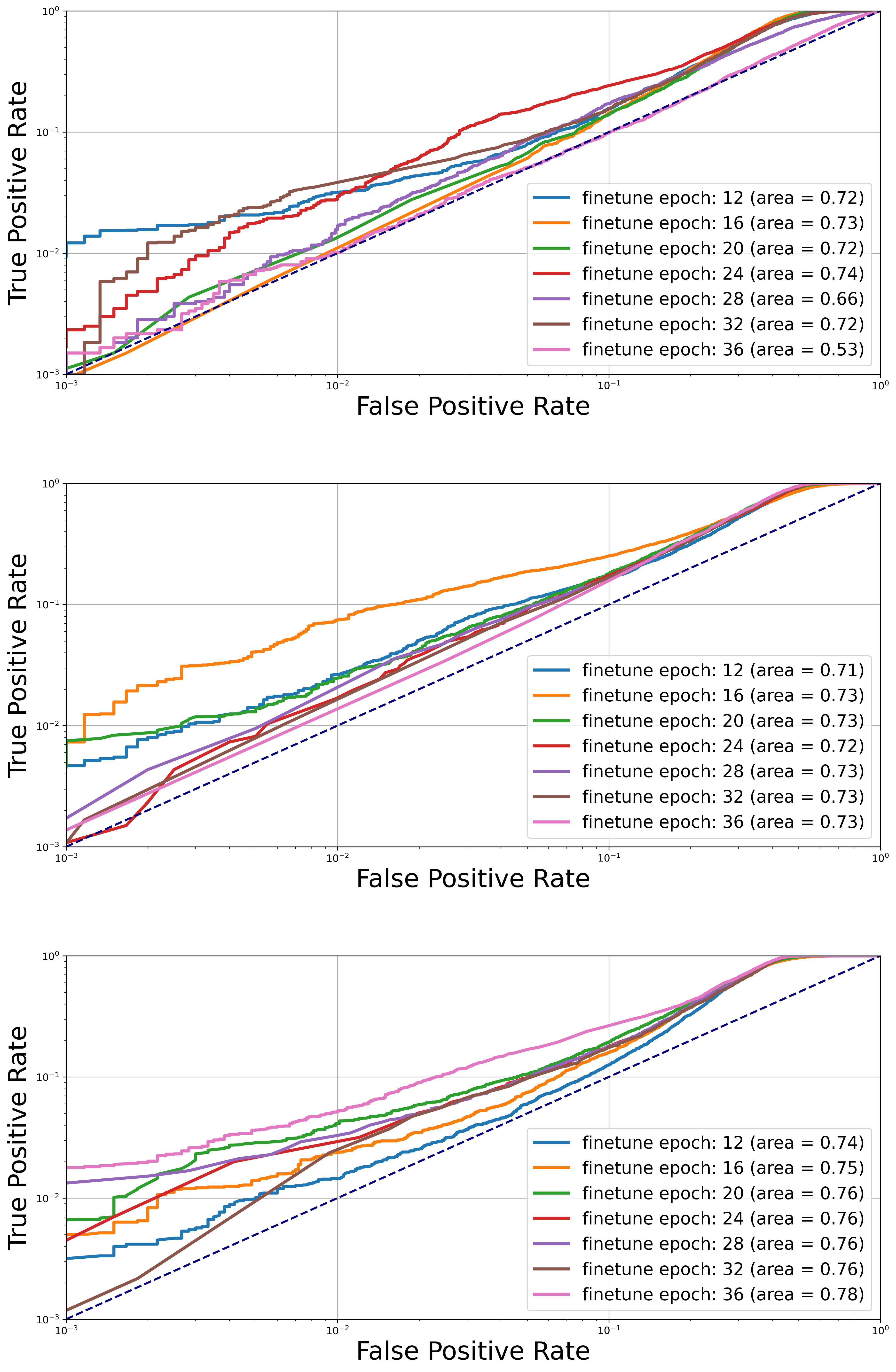}
                 \caption{backward properties}
                 \label{fig:backward information}
             \end{subfigure}
                \caption{Log-scale ROC curves for the audit across three model sizes on PubMed-RCT dataset at various fine-tuning epochs (from top to bottom: GPT-2-medium, GPT-2-large, GPT-2-xl)}
                \label{fig:two graphs}
        \end{figure}

        \begin{figure}
            \centering
            \begin{minipage}{0.48\textwidth}
                \centering
                \includegraphics[width=\textwidth]{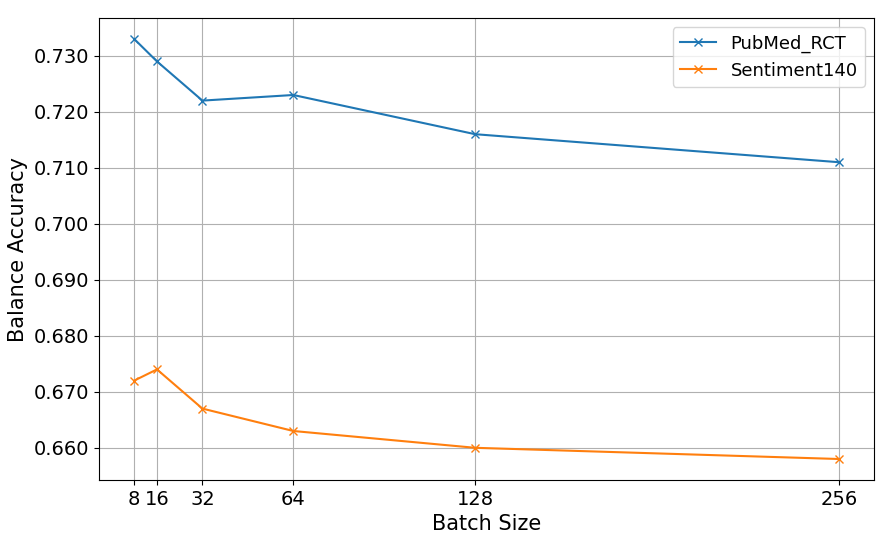}
                \caption{The balance audit accuracy with different batchsizes for GPT-2-medium trained on PubMed\_RCT and Sentiment140.}
            \end{minipage}
            \hspace{0.01\textwidth}
            \begin{minipage}{0.47\textwidth}
                \centering
                \includegraphics[width=\textwidth]{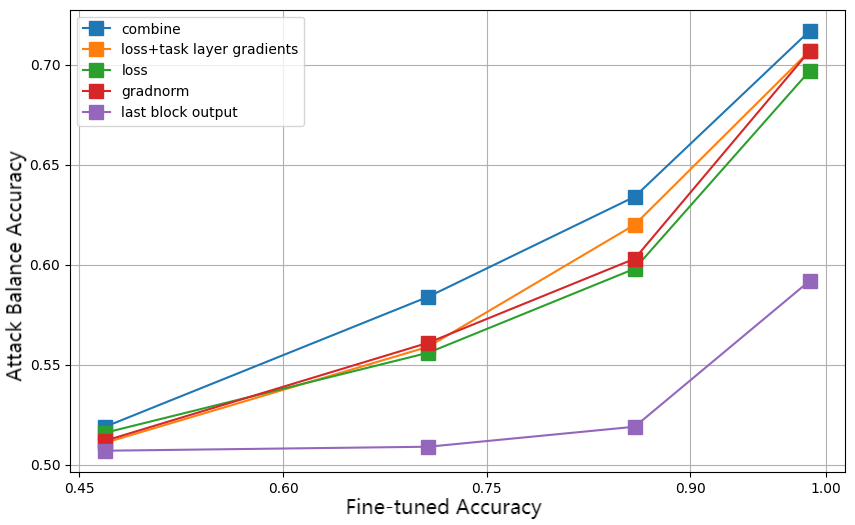}
                \caption{The trade-off between fine-tuning accuracy and attack accuracy with different information for GPT-2-medium trained on PubMed\_RCT.}
            \end{minipage}
        \end{figure}

        \textbf{Model Size and Different Properties.} To investigate the effects of model size and various properties on privacy audit during the fine-tuning of LMs, we performed additional experiments using the medium, large, and xl versions of GPT-2. As shown in Figures 3 and 4, the models from top to bottom are medium, large, and xl, respectively.
        
        \begin{itemize}
            \item Larger models pose a higher privacy leakage risk. In our experiments, the medium model's forward attack accuracy was 62.7\%, and backward attack accuracy was 74.3\%. For the XL model, these figures were 67.8\% and 75.1\%, respectively. This trend is evident in the ROC curves and AUC values on a logarithmic scale. 
            \item Forward attacks are weaker than backward attacks in accuracy. For GPT-2-large, backward attacks reach 73.9\% accuracy, while forward attacks peak at 66.4\%. However, in TPR at low FPR, forward attacks perform better, and show no significant difference in AUC.
        \end{itemize}
        
        The main reason is that models with a larger number of parameters are more likely to overfit on small datasets, which is crucial for effective MIA. Furthermore, larger models tend to generate more unique and specific outputs, increasing the risk of privacy breaches during audits.
        
        The information used by the method includes loss and gradients, which are directly related to how the model adjusts to specific data during training. In contrast, forward properties only provide the model's response to specific inputs. While backward properties offer attackers richer and deeper insights, it is highly asymmetric for privacy auditing. After training, data from the same dataset typically show low loss and nearly identical gradient representations. Consequently, under the influence of backward information, the inference model is more inclined to predict these as member samples. This is why inference models utilizing forward and backward information exhibit entirely different patterns across various metrics. Therefore, the contributions of forward and backward information to the inference model vary, necessitating better integration to enhance the model's inference performance.
        
        \textbf{Batch Size and Training Iterations.} Apart from specific text tasks, text lengths, model sizes, and training iterations, during the experiments we also discovered that the batch size set for fine-tuning training impacts the values of the privacy leakage audit metrics, as shown in Figure 5. Larger fine-tuned batch sizes during training, result in lower privacy leakage risks.
        
        \textbf{The Trade-off Between Privacy and Utility.} Privacy auditing reveals models' privacy risks and helps trainers balance performance with privacy, aiming for optimal efficacy with minimal data exposure. Figure 6 shows that as model accuracy improves during GPT-2 training on RCT data, attack success rates also increase. It is evident that when the fine-tuning accuracy reaches 82.5\%, the highest balance accuracy of the attack is 71.7\%. Conversely, at a fine-tuning accuracy of 70.7\%, the highest balance attack accuracy is 58.4\%, with a forward information attack success rate of 50.9\%. This indicates the risk of privacy data leakage even before the model shows signs of overfitting. 
        
        As shown in Figure 3, after a single epoch of fine-tuning, the auditing technique identified privacy risks in the dataset with a success rate of 55\%, even though the model loss had not yet reached its lowest value, indicating that the model could be vulnerable to malicious privacy attacks even before it is overtrained. In addition, irrespective of the information that is chosen, the audit metrics initially peaks and then steadily decreases as the target model undergoes further fine-tuning. This suggests that overtraining does not inherently increase privacy risks. One possible explanation is that the target model first overfits certain aspects of the fine-tuning data. As training progresses, the extent of overfitting on these specific aspects grows, but this does not mean that more aspects are overfitted, and it may even show a decreasing trend. Overfitting does not always appear in the model's output behavior; loss is just one dimension of overfitting.
        
        To achieve a balance between privacy and utility without modifying the model, an effective strategy is to set a predefined maximum attack accuracy and a minimum required model accuracy. However, to determine the best trade-off point, it is essential to devise more robust metrics. 

        \textbf{Extension of Parameter-efficient Fine-tuning.} Our work provides a comprehensive quantification of the privacy risks associated with full-parameter fine-tuning (FFT). However, it is also necessary to consider the impact of parameter-efficient fine-tuning (PEFT) on model privacy leakage. FFT requires significant parameter updates, incurs high computational costs, and does not necessarily guarantee better performance, making PEFT, which fine-tunes only a small subset of parameters, a more popular choice currently. Taking real-world applications into account, we have supplemented our framework with experiments on PEFT.
        \begin{table*}
            \centering
            \setlength{\tabcolsep}{2.5mm}
            \fontsize{10pt}{14pt}\selectfont
            \begin{tabular}{l|cccc} 
                \toprule
                PEFT & \# Parameters & PubMed\_RCT & Yelp Reviews & PubMedQA\\
                \midrule
                FFT & 6.7b & 0.766 & 0.733 & 0.773\\
                LoRA & 33.6M & 0.672 & 0.639 & 0.683\\
                Prefix Tuning  & 5.3M & 0.654 & 0.614 & 0.663\\
                $(IA)^3$ & 0.6M & 0.627 & 0.6 & 0.641\\
                \bottomrule
            \end{tabular}
            \caption{The \textit{balance accuracy} performance of \textsc{Parsing} across Llama2-7b fine-tuned with different PEFT techniques (from top to bottom: FFT, LoRA, Prefix Tuning, $(IA)^3$) over three datasets.}
        \end{table*}
        \begin{table*}
            \centering
            \setlength{\tabcolsep}{2.5mm}
            \fontsize{10pt}{14pt}\selectfont
            \begin{tabular}{l|cccc} 
                \toprule
                Model & Methods & \# Parameters & Balance accuracy & TPR at 0.1 FPR \\
                \midrule
                GPT-2-medium & FFT & 345M & 0.701 & 0.335\\
                Llama2-7b & FFT & 6.7B & 0.766 & 0.403\\
                Llama2-7b & LoRA & 33.6M & 0.672 & 0.304\\
                Llama2-13b & LoRA & 65.5M & 0.679 & 0.317\\
                \bottomrule
            \end{tabular}
            \caption{The \textit{balance accuracy} and \textit{TPR at 0.1 FPR} performance of \textsc{Parsing} across Llama2-7b and  Llama2-13b models fine-tuned with LoRA techniques over PubMed\_RCT dataset.}
        \end{table*}
        
        The experiments were conducted on the Llama2-7b and Llama2-13b models to better adapt to the application scenarios of PEFT. The fine-tuning datasets included the text classification dataset \textit{PubMed\_RCT}, the sentiment analysis dataset \textit{Yelp Reviews}, and the QA dataset \textit{PubMedQA}. The data partitioning for these datasets follows the settings in Table 3. The evaluation metric used were \textit{Balance Accuracy} and \textit{TPR at 0.1 FPR}, to provide a clearer comparison of the privacy risk differences between various fine-tuning methods. The training parameters for both the target model and the audit model also follow the settings in Table 5. In these experiments, for LoRA fine-tuning, setting the rank to 8 and fine-tuning the q\_proj and v\_proj modules. The experimental results are shown in Tables 3 and 4.
        
        From the experimental results, it can be seen that using PEFT can reduce the privacy risk of the fine-tuning dataset without significantly degrading model performance. Additionally, the attack success rate decreases as the number of fine-tuned parameters decreases. For example, for the \textit{PubMed\_RCT}, the balance accuracy under FFT is approximately 0.766. When the number of tunable parameters is reduced to 33.6M, the balance accuracy drops to 0.672. When the number of training parameters is further reduced to below 1M, the balance accuracy drops to just 0.627. This is an important finding, as it at least validates that the amount of tunable parameters in model fine-tuning is a critical factor in protecting specific LM fine-tuning from MIA. 
        
        We analyze the reason for this: in FFT, all the model's parameters are adjusted, which may cause the model to more easily memorize specific samples from the training data. This comprehensive retention of training data by the LLM increases the likelihood of successful MIA. Moreover, by extending the audit framework to larger 13b models, the framework successfully executed the audit, revealing the risks of privacy leakage in fine-tuned models. 
        
        \textbf{Others.} Additional experiments and discussions, such as experiments on larger models, information redundancy of different properties, other factors affecting the privacy vulnerabilities of LMs fine-tuning, and privacy mitigation strategies, are all provided with experimental results and detailed analyses in the appendix.
        
        \section{Conclusion}
        Our research provides valuable insights and tools for enhancing privacy resilience in the fine-tuning of LMs. By introducing a robust privacy auditing framework \textsc{Parsing}, demonstrating its efficacy through rigorous experiments, and thoroughly analyzing the privacy characteristics of LMs fine-tuning, we contribute to the development of safer and more reliable NLP applications. Future work will focus on extending the framework to more realistic fine-tuning applications and dynamic learning environments, with the aim of further refining privacy protection mechanisms in the evolving landscape of LM applications.

	\bibliographystyle{plainnat}

	\newpage
	\appendix
	
	\maketitle
        \section{Experimental Dataset and Target Model}
        In this study, we fine-tuned LMs for various text tasks, such as topic classification and knowledge-based question answering, and perform a privacy audit of the model fine-tuning process under an audit framework. To showcase the framework's task adaptability and model adaptability, we selected several specific, typical text task datasets and pre-trained language models. Datasets specifically include PubMed\_RCT, Yelp Reviews, Wiki Toxicity, and so on. Table 5 provides details on the text tasks associated with each dataset, their rough original data volumes, and the characteristics of the text length. Table 6 details the pre-trained models used and their sizes.
        \begin{table}[b]
            \centering
            \setlength{\tabcolsep}{5mm}
            \fontsize{10pt}{14pt}\selectfont
            \begin{tabular}{lcccc} 
                \toprule
                Datasets & Text Tasks & Data Volume & Text Length \\
                \midrule
                PubMed\_RCT & Topic C. & 180k+60k & long\\
                Yelp Reviews & Sentiment A. & 5200k & short\&long\\
                BC5CDR & NER. & 1500(article) & long \\
                PubMedQA & QA. & 210k & short\&long \\
                Wiki Toxicity & Content R. & 100k+20k & short\&long\\
                AG News & Topic C. & 120k & short \\
                Sentiment140 & Sentiment A. & 1,600k & short \\
                CoNLL-2003 & NER. & 15k+7k & short\\
                \bottomrule
            \end{tabular}
            \caption{Description of the datasets used in the research.}
        \end{table}
        
        \begin{table}[t]
            \centering
            \setlength{\tabcolsep}{6mm}
            \fontsize{10pt}{14pt}\selectfont
            \begin{tabular}{lcccc} 
                \toprule
                Models & Layers & Params\\
                \midrule
                GPT-2-medium & 24 & 345M\\
                GPT-2-large & 36 & 774M\\
                GPT-2-xl & 48 & 1.5B\\
                GPT-Neo & 24 & 1.3B\\
                GPT-Neo & 32 & 2.7B\\
                Llama2-7b & 32 & 7B\\
                Llama2-13b & 40 & 13B\\
                \bottomrule
            \end{tabular}
            \caption{Description of the target base model used in the research.}
        \end{table}

        \section{Dataset Partition and Setup}
        \label{sec:appendix}
        \begin{table*}
            \centering
            \setlength{\tabcolsep}{2mm}
            \fontsize{10pt}{14pt}\selectfont
            \begin{tabular}{l|cc|cccc} 
                \toprule
                \multirow{2}{*}{Datasets}& \multicolumn{2}{c|}{Target Model} & \multicolumn{4}{c}{Auditing Model}\\
                \cline{2-7} 
                & training & test & training members & training non-members & test members & test non-members \\
                \midrule
                PubMed\_RCT & 20000 & 20000 & 5000 & 5000 & 6000 & 6000\\
                Yelp Reviews & 20000 & 20000 & 5000 & 5000 & 5000 & 5000\\
                BC5CDR & 20000 & 10000 & 5000 & 5000 & 2500 & 2500\\
                PubMedQA & 20000 & 20000 & 5000 & 5000 & 5000 & 5000\\
                Wiki Toxicity & 20000 & 20000 & 5000 & 5000 & 5000 & 5000\\
                AG News & 20000 & 20000 & 5000 & 5000 & 5000 & 5000\\
                Sentiment140 & 20000 & 20000 & 5000 & 5000 & 5000 & 5000\\
                CoNLL-2003 & 10000 & 10000 & 5000 & 5000 & 2500 & 2500\\
                \bottomrule
            \end{tabular}
            \caption{Size of datasets used for training and testing the target LM model and the auditing model.}
        \end{table*}
        
        In this study, we fine-tuned the LMs on various text tasks and assessed our privacy audit framework during the fine-tuning process. These tasks included topic classification, sentiment analysis, named entity recognition, knowledge-based question answering, and content review. We selected representative and suitable data sets for each task. In these datasets, we not only uncovered the significant privacy risks during the LM fine-tuning process, but also reached some intriguing conclusions.
        
        However, because of the substantial differences in the volumes of data between the different datasets, we further subdivided each dataset to eliminate experimental interference as much as possible, with the aim of reaching more scientific conclusions. Table 7 illustrates the sizes of the datasets used in training and testing of both the target model and the audit model. For the target model, \textit{training} indicates the volume of data used for training the target model, and \textit{test} denotes the size of the dataset used for testing the target model. \textit{Training members} refers to the member samples used for training the audit model, which belong to the target model's training set; \textit{Training non-members} indicates the non-member samples used for training the audit model, which do not belong to the target model's training set; \textit{Test members} refers to the member samples used for testing the audit model; while \textit{test non-members} denotes the non-member samples used for testing the audit model.
        
        During the division process, we ensured that the training and test sets of each data set were balanced to maintain the reliability of the evaluation and statistical significance. The balanced division of data sets for both the target model and the audit model enables comparisons of the audit model under the same conditions. By clearly differentiating between member and non-member samples, it is possible to fine-tune the evaluation if the model has memorized the details of the training data (i.e., member samples), thus more precisely quantifying the privacy leakage risk. Furthermore, setting a fixed data amount for each training set of the audit model (e.g., 5000 member samples and 5000 non-member samples) helps to control the experiment's scale and computational complexity, enhancing the comparability and reproducibility of the experimental results.

        \section{Additional Argumentation}
        \label{sec:appendix}
        \subsection{Notations}
        \begin{table*}[!t]
            \centering
            \setlength{\tabcolsep}{2mm}
            \fontsize{10pt}{12pt}\selectfont
            \begin{tabular}{ccc}
                \toprule
                Symbol & Description & Range \\
                \midrule
                $x$ & Input sample of the target fine-tuned model or audit model. & - \\
                $y$ & Prediction of the target fine-tuned model. & - \\
                $M_{ft}$ & Target fine-tuned model. & - \\
                $M_{pre}$ & Pre-trained model. & - \\
                $\mathcal{M}$ & Intermediate module of the target fine-tuned model. & - \\
                $L$ & Total number of intermediate modules in the target fine-tuned model. & - \\
                $l$ & Intermediate module ID. & $1-L$ \\
                $\boldsymbol{\theta}_{ft}$ & Parameters of the fine-tuned model. & - \\
                $\mathcal{L}$ & Target model loss for a specific input $x$. & - \\
                $\mathcal{K}$ & Knowledge set accessible to the auditor. & - \\
                $D^{ft}_{train}$, $D^{ft}_{test}$ & Fine-tuning dataset. & - \\
                $D^{inf}_{train}$, $D^{inf}_{test}$ & Audit dataset. & - \\
                $N$ & Total number of samples in $D^{ft}_{train}$. & - \\
                $\alpha$ & The scaling factor. & $0-1$ \\
                $\mathcal{I}$ & The white-box candidate property repository. & - \\
                $\nabla$ & Gradient operator. & - \\
                $r(x)$ & The representation embedding for a sample $x$. & - \\
                $\mathcal{F}$ & The property alignment process. & - \\
                $\mathbf{R}$ & The property embedding generator. & - \\
                $wf$, $wb$ &  The learnable parameters associated with the forward and backward embedding generators. & - \\
                $\oplus$ & The concatenation operation. & - \\
                $\lambda$ & The parameter for information weighting. & $0-1$ \\
                $\mathcal{S}$ & The set of sample embeddings. & - \\
                $L$ & Learning objective of the property embedding. & - \\
                $\mu(t)$ & A weight function. & $0-1$ \\
                $k$ & The decay coefficient. & - \\
                $m$ & A boundary parameter. & - \\
                $d$ & The cosine similarity between two vectors. & - \\
                \bottomrule
            \end{tabular}
            \caption{Notation table of the math symbols and the description.}
        \end{table*}
        The methodology section of this paper involves multiple mathematical symbols and notations, covering key concepts related to mathematical computations, models, and datasets. Additionally, some custom symbols are included to more accurately describe the proposed algorithms and experimental steps. For the reader's convenience in referencing and understanding the specific meanings of these symbols, please refer to the notation table, which provides detailed explanations of the definitions and ranges of these symbols. This will help in better understanding the technical details and mathematical derivations presented in this paper.
    
        \subsection{Further Argumentation of the Method}
        In Section 4, we present a detailed overview of the principles and workflow of \textsc{Parsing}. Here, we provide further argumentation. First, for the auditing model training set $D^{inf}_{train}$ obtained during data partitioning, we reshape it into nested pairs$([x_0, y_0, 1], [x_1, y_1, 0])$, where $(\cdot)$ represents a set of samples. Each element $[\cdot]$ includes the input text data, the task label and the membership label. The membership label of 1 indicates that the sample belongs to the fine-tuning training set $D^{inf}_{train}$ of the target model, while 0 indicates that the sample belongs to a mutually exclusive data set with the same distribution as $D^{inf}_{train}$, usually sampled from the same parent data set.

        \begin{figure}[!t]
          \centering
          \includegraphics[width=0.7\linewidth]{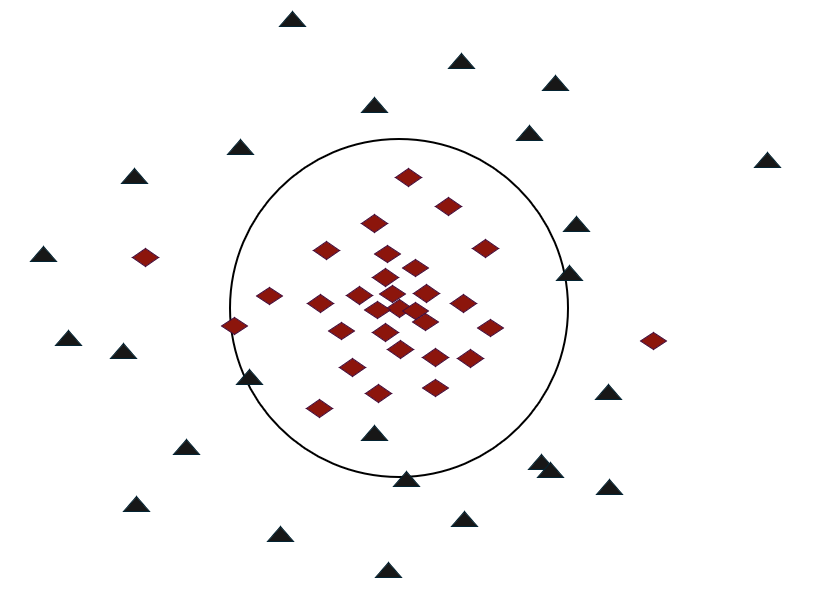}
          \caption{Schematic representation of learning objectives.}
        \end{figure}

        During audit model training, the embedded membership representation for all samples in the same batch is calculated using Equation (6). For each set of samples, we introduced the distance metric described in Equation (7) to differentiate between member and non-member samples. It is important to further clarify that for $L_{d}$, within each group in a batch, we compute the representation difference between member and non-member samples, while $L_{s}$ calculates the concentration loss of member samples. $L_{s}$ primary purpose is to ensure that the representations of member samples in the embedding space are as similar as possible, thereby improving the accuracy of member inference.

        In practice, we typically care more about samples predicted to be members, or rather, we aim to find the internal similarity of member samples. The internal representation of non-member samples is less of a concern, as long as their interclass distance from member samples is sufficiently large. In privacy audit tasks, similar to the concept of single-class clustering as illustrated in Figure 7, we do not assign equal attention to the samples predicted as 0 and 1.
        
        The learning objective introduced in this method does not require carefully designed pairs or triplets, but merely ensures that each data set conforms to a predefined form, simplifying computation while enhancing training stability. Additionally, by applying concentration loss to the member sample set, it improves the clustering effect of member sample representations, ensures the consistency of member samples, and reduces the impact of noise on the audit model.

        Combine the losses of the two components for joint optimization, The final optimization objective combines the aforementioned losses:
        \begin{equation}
          L_{total}=L+\nu L_{ce},
        \end{equation} 
        where $\nu$ represents the weight parameter, balance the influence of each part's loss.  

        The time complexity for data partitioning and preprocessing is $O(N)$, where $N$ denotes the number of samples. The time complexity for computing embedding representations depends on the structure of the embedding representation generator (full convolutional neural network). Let the time complexity for the forward and backward information embedding representation of a single sample be $O(f(n))$ and $O(b(n))$, respectively. The overall time complexity for the calculation of the embedding is $O(N\cdot(f(n)+b(n))$. The complexity of calculating the representation distance metric is $O(N\cdot d)$, where $d$ represents the complexity of calculating the representation distance. The total time complexity of the concentration loss of the set of sample members is $O(N^2\cdot d)$. Finally, the time complexity for the classifier to compute the cross-entropy loss between each sample's predicted label and the true label is $O(N)$. If the count of model parameters is p, the time complexity for each gradient update is $O(p)$, and the total time complexity of the algorithm is $O(T\cdot(N\cdot d+N^2\cdot d+p))$, where $T$ is the number of iterations. In fact, the primary time consumption comes from gradient extraction. In our tests, for GPT-medium, extracting all layer gradients for conducting a backpropagation-based audit using four NVIDIA A100 40GB GPUs with a batch size set to 64 takes approximately 4 seconds to process one batch, with peak memory usage around 12GB/GPU.

        \subsection{Pseudocode}
        \begin{algorithm*}
        \caption{\small PARSING: The Detailed Privacy Auditing Procedure for Supervised Fine-tuning of Language Models}
        \label{alg:detailed_parsing_algorithm}
        \textbf{\small Input}:\small Pre-trained model $M_{\text{pre}}$ with parameters $\boldsymbol{\theta}_{\text{pre}}$, Fine-tuning dataset $D_{\text{ft}}$\\
        \textbf{\small Parameter}:\small Epochs $\mathcal{N}$, Scaling factor $\alpha$, Alignment parameter $\lambda$, Boundary parameter $m$,\\ Decay coefficient $k$, Audit interval $epoch_{\text{interval}}$\\
        \textbf{\small Output}:\small Membership Inference Module $M_{\text{inf}}$, Auditing metrics (e.g., AUC, TPR at 0.1 FPR)

        \small 
        \setstretch{0.9}
        \begin{algorithmic}[1] %[1] enables line numbers
        \STATE Initialize fine-tuning model $M_{\text{ft}}$ with pre-trained parameters $\boldsymbol{\theta}_{\text{pre}}$
        \STATE Split fine-tuning dataset $D_{\text{ft}}$ into training set $D_{\text{ft}}^{\text{train}}$ and test set $D_{\text{ft}}^{\text{test}}$.
        
        \STATE \textbf{\# Step 1: Data Partitioning:}
        \STATE Randomly partition $D_{\text{ft}}^{\text{train}}$ and $D_{\text{ft}}^{\text{test}}$ into disjoint subsets:
        \STATE $D_{\text{inf}}^1 = \{(x_j, y_j, 1) \mid j = 1, 2, ..., \alpha N\}$ \COMMENT{Members, labeled as 1} 
        \STATE $D_{\text{inf}}^0 = \{(x_k, y_k, 0) \mid k = 1, 2, ..., \alpha N\}$ \COMMENT{Non-members, labeled as 0}
        \STATE Construct training set $D_{\text{inf}}^{\text{train}} = D_{\text{inf}}^1 \cup D_{\text{inf}}^0$
        \STATE Initialize membership inference module $M_{\text{inf}}$
        \STATE \textbf{Batch Processing:} Reshape the batch $\mathcal{B}$ into nested pairs $([x_0, y_0, 1], [x_1, y_1, 0])$
        
        \FOR{epoch $i=1$ to $\mathcal{N}$}
            \STATE Fine-tune $M_{\text{ft}}$ on $D_{\text{ft}}^{\text{train}}$
            \IF{$i \bmod epoch_{\text{interval}} = 0$} 
                \FOR{each batch $\mathcal{B}$}
                \FOR{each sample $(x, y)$ in batch $\mathcal{B}$}
                    
                    \STATE \textbf{\# Step 2: Property Extraction}
                    \IF{Forward properties are required}
                        \STATE Extract forward properties $\mathcal{I}_f$:
                        \FOR{each intermediate layer $l$ in $M_{\text{ft}}$}
                            \STATE Compute $x_{l+1} = \mathcal{M}_l(x_l)$ \COMMENT{Output of intermediate layer $l$}
                        \ENDFOR
                        \STATE Store forward properties $\mathcal{I}_f = \{\mathcal{M}_l(x_l) \mid 1 \leq l \leq L\} \cup \{x, y\}$
                    \ENDIF
                    
                    \IF{Backward properties are required}
                        \STATE Extract backward properties $\mathcal{I}_b$:
                        \STATE Compute loss $\mathcal{L}(M_{\text{ft}}(x), y)$
                        \STATE Compute gradients $\nabla_{\boldsymbol{\theta}}\mathcal{L}(M_{\text{ft}}(x), y)$
                        \STATE Compute gradient norms $\|\nabla_{\boldsymbol{\theta}}\mathcal{L}\|_p$
                        \STATE Store backward properties $\mathcal{I}_b = \{\nabla_{\boldsymbol{\theta}}\mathcal{L}(M_{\text{ft}}(x), y), \|\nabla_{\boldsymbol{\theta}}\mathcal{L}\|_p, \mathcal{L}(M_{\text{ft}}(x), y)\}$
                    \ENDIF
                \ENDFOR

                    \STATE \textbf{\# Step 3: Property Embedding}
                    \IF{Forward properties are selected}
                        \STATE Normalize and align forward properties $\mathcal{F}_f(\mathcal{I}_f)$
                        \STATE Generate forward property embedding $\textbf{R}_{wf}[\mathcal{F}_f(\mathcal{I}_f)]$
                    \ENDIF
                    
                    \IF{Backward properties are selected}
                        \STATE Normalize and align backward properties $\mathcal{F}_b(\mathcal{I}_b)$
                        
                        \STATE Generate backward property embedding $\textbf{R}_{wb}[\mathcal{F}_b(\mathcal{I}_b)]$
                    \ENDIF
                    
                    \STATE Combine embeddings: $r(x) = \lambda \textbf{R}_{wf}[\mathcal{F}_f(\mathcal{I}_f)] \oplus (1-\lambda) \textbf{R}_{wb}[\mathcal{F}_b(\mathcal{I}_b)]$
                
                    \STATE Compute \textit{embedding difference loss} $L_{d} = \frac{1}{|\mathcal{B}|}\sum_{i=1}^{|\mathcal{B}|} \max \left[0,m-d \left(r_{\text{mem},i},r_{\text{non},i} \right)\right]$
                    \STATE Compute \textit{embedding concentration loss} $L_{s} = \frac{1}{|\mathcal{B}|}\sum_{i,j=1}^{|\mathcal{B}|} d'(r_{\text{mem},i}, r_{\text{mem},j})$
                    \STATE Combine the losses $L = L_{diff} + \lambda L_{sim}$
        
                    \STATE \textbf{\# Step 4: Membership Inference} 
                    \STATE Compute Classifier Training loss $L_{ce} = -\frac{1}{N}\sum_{i} [m_i \log(\hat{m}_i) + (1 - m_i) \log(1 - \hat{m}_i)]$
                    \STATE Joint Optimization: Combine the total loss $L_{\text{total}} = L + \nu L_{ce}$ and perform backpropagation to update model parameters.
                \ENDFOR
                \STATE \textbf{\# Step 5: Evaluation}
                \STATE Compute auditing metrics (e.g., balanced accuracy, AUC, TPR at low FPR) based on the performance of $M_{\text{inf}}$
            \ENDIF
        \ENDFOR
        
        \STATE \textbf{return} Trained membership inference module $M_{\text{inf}}$, final auditing metrics
        \end{algorithmic}
        \end{algorithm*}

        \textsc{Parsing} merges metric-based and classification-base methods, eschewing reliance on the target model's output accuracy for classification. It focuses on representation learning from white-box properties of various samples to create distinct member embeddings. Learning then targets the distances between these embeddings, enabling the inference model to discern the differences in information representation between member and non-member samples within the target model. In the main text, we provide a detailed introduction to the principles and workflow of the method framework. Algorithm \ref{alg:detailed_parsing_algorithm} offers a step-by-step breakdown of the entire process, including data partitioning, property extraction, embedding generation, and membership inference, culminating in the joint optimization of loss functions to achieve accurate member classification.

        \section{Detailed Experimental Setup }
        \label{sec:appendix}
        \begin{table*}
            \centering
            \setlength{\tabcolsep}{3mm}
            \fontsize{10pt}{14pt}\selectfont
            \begin{tabular}{lccc|ccccc} 
                \toprule
                \multirow{2}{*}{Tasks} & \multicolumn{3}{c|}{Target Model} & \multicolumn{5}{c}{Audit Model}\\
                \cline{2-9} 
                & leaning rate & batch size &  max-seq-len & $\mu_0$ & $k$ & $\lambda$ & $\nu$ & learning rate \\
                \midrule
                PubMed\_RCT & 2e-5 & 128/64/32 & 128 & 0.1-0.8 & 5e-3 & 0.2-0.5 & 1 & 5e-4\\
                Yelp Reviews & 2e-5 & 128/64/32 & 128 & 0.1-0.8 & 5e-3 & 0.2-0.5 & 1 & 5e-4\\
                BC5CDR & 2e-5 & 64/32/16 & 128 & 0.1-0.8 & 5e-3 & 0.2-0.5 & 1 & 5e-4\\
                PubMedQA & 1e-5 & 64/32/16 & 128 & 0.1-0.8 & 5e-3 & 0.2-0.5  & 1 & 5e-4\\
                Wiki Toxicity & 2e-5 & 128/64/32 & 128 & 0.1-0.8 & 5e-3 & 0.2-0.5  & 1 & 5e-4\\
                AG News & 2e-5 & 128/64/32 & 128 & 0.1-0.8 & 5e-3 & 0.2-0.5  & 1 & 5e-4\\
                Sentiment140 & 2e-5 & 128/64/32 & 128 & 0.1-0.8 & 5e-3 & 0.2-0.5  & 1 & 5e-4\\
                CoNLL-2003 & 2e-5 & 128/64/32 & 128 & 0.1-0.8 & 5e-3 & 0.2-0.5  & 1 & 5e-4\\
                \bottomrule
                \end{tabular}
            \caption{Main hyperparameters and experimental settings used in training and testing the target model and audit model.}
        \end{table*}
        Table 9 presents the main hyperparameters and experimental settings used in training and testing the target model and audit model. We endeavored to maintain consistency in hyperparameters across different tasks; however, due to computational resource constraints, we adjusted certain hyperparameter values when fine-tuning specific tasks on larger models.

        The computational configuration for the experiment included four NVIDIA A100 40GB GPUs. Using data parallelism technology, the memory demand was distributed across multiple GPUs, ensuring sufficient memory while accelerating training speed and providing redundancy. All audit models were trained for 8 epochs, with additional training rounds on certain tasks for better audit results. Each set of experiments was repeated 10 times, and the average and standard deviation of the calculated metrics were used as a measure of the statistical distribution.

        For $\textsc{Parsing}$, the white-box information used of the target model includes:
        \begin{itemize}
            \item Loss - the cross-entropy loss or possibly other losses.
            \item Task labels - the true labels of the data samples.
            \item Outputs - the output of the intermediate modules.
            \item Predicted labels - the predicted labels of the task layer.	
            \item Gradients - the gradients of the last attention module.	
            \item Gradient norms - the gradient norms of all layers.	
        \end{itemize}

        \section{Validation of Information Redundancy}
        \label{sec:appendix}
        To gain a deeper understanding and analysis of how different combinations of properties affect inference performance, we used combinations of output information from different layers to feed into the audit model for separate attacks. In attacks based on forward properties, we found that the accuracy of attacks using only the output information of the last layer (excluding any additional task-specific layers) was always higher than those using output information from other layers or their combinations. This indicates that the output of the last layer leaks the most information about the members of the training data. 
        
        This experimental phenomenon can be explained: Lower-level modules extract simple features from the input, and these simple features exhibit very similar patterns across all data from the same distribution, making it difficult to optimize distinguishable membership property embeddings. In contrast, higher-level modules extract more abstract features from the input, which carry more complex membership information. Higher-level modules also pay more attention to non-task-specific features (noise) in the training data, making it easier to optimize and generate superior membership property embeddings. Therefore, using only the output information from the last layer can achieve better membership inference accuracy.
        
        Thus, we considered another question: In membership inference for fine-tuned models, is the information provided by other layers entirely redundant? Table 10 presents the overlap in the number of correctly inferred samples by the audit model under different combinations of forward properties during the fine-tuning of PubMed\_RCT on GPT-2-xl. \textit{Information} refers to the combinations of output information from different layers. \textit{Correct Prediction} indicates the number of correctly predicted samples. \textit{Overlapping Samples} represent the number of overlapping member and non-member samples among all correctly predicted samples, compared to *. It is clear that the correctly predicted samples in each round of auditing do not form a subset relationship. Therefore, the information provided by the different layers of the model is not entirely redundant, indicating a need for further research and exploration.
        \begin{table*}
            \centering
            \setlength{\tabcolsep}{3mm}
            \fontsize{10pt}{14pt}\selectfont
            \begin{tabular}{cccccc} 
                \toprule
                \multirow{2}{*}{Information}& \multirow{2}{*}{Sample Size} & \multirow{2}{*}{Correct Prediction} & \multicolumn{3}{c}{Overlapping Samples}\\
                \cline{4-6} 
                &  &  & non-members & members & total\\
                \midrule
                Last layer & 10000 & 6858* & / & / & / \\
                Last 2 layers & 10000 & 6656 & 2088 & 2740 & 4828\\
                Last 4 layers & 10000 & 6607 & 2047 & 2631 & 4678\\
                Last 8 layers & 10000 & 6615 & 1924 & 2717 & 4641\\
                Last 16 layers & 10000 & 6623 & 2165 & 2492 & 4657\\
                \bottomrule
                \end{tabular}
            \caption{Under different combinations of forward properties, the overlap in the number of correctly inferred samples by the audit model on the same test dataset.}
        \end{table*}

        \section{More Experiments and Results}
        \label{sec:appendix}
        \subsection{Extension of More Target Models}
        We conducted extensive validation experiments using open-source pre-trained models of various sizes to support our discussion in the main text. The detailed experimental results are presented in Tables 11 to 20. Specifically, Tables 11, 12, 16, and 17 correspond to experiments conducted on GPT-2, while Tables 13, 14, 18, and 19 pertain to GPT-Neo. Experiments in Tables 15 and 20 were carried out on Llama2-7b. Tables 11 to 15 report the maximum values for \textit{balanced accuracy} and \textit{AUC} during model fine-tuning on different text tasks. Meanwhile, Tables 16 to 20 list the maximum values for \textit{TPR at 0.1 FPR} under similar conditions. The statistical precision of AUC is 0.005, while the statistical precision of the balance accuracy and $TRR_{0.1}$ is 0.001.
        
        Apart from the findings in the main text, as seen in Tables 13 and 18, GPT-Neo-1.3b has generally lower privacy audit metrics during the fine-tuning process compared to GPT-2-XL. For example, for the BC5CDR task, the risk of privacy leakage for GPT-Neo-1.3b is $ balance\_acc=.733$; $auc=.77$; $TPR_{0.1}=0.368$, while it is $ balance\_acc=.744$; $auc=.775$; $TPR_{0.1}=0.334$ for GPT-2-XL. However, there is not much difference in the parameter sizes of the two models. We explain the reasons as follows: on the one hand, GPT-Neo has performed multiple architectural optimizations to enhance training efficiency and performance, indirectly improving the model's generalization capability. On the other hand, with nearly equal parameter counts, GPT-Neo has only half the number of layers as GPT-2, and in our experiments, we set the same maximum token sequence length. Thus, to a certain extent, GPT-Neo is better at handling membership inference attacks under the same parameter conditions compared to GPT-2.
        
        Furthermore, although Llama2-7b reaches a higher maximum privacy risk during the audit process compared to other smaller models, this maximum value appears very early, with privacy being most vulnerable in the early stages of fine-tuning, then decreasing and converging to a value. One possible reason is that, relative to the limited fine-tuning data, the model's parameter size is too large, causing it to overfit a large amount of data early in the training. However, as training continues, the model increasingly fits the noise and specific details of some data, while relaxing its ``grip" on other more reasonably distributed data, resulting in the gradual convergence of privacy audit outcomes.

        \begin{table*}
            \centering
            \setlength{\tabcolsep}{2mm}
            \fontsize{10pt}{14pt}\selectfont
            \begin{tabular}{l|ccc|ccc} 
                \toprule
                \multirow{2}{*}{Datasets}& \multicolumn{3}{c|}{Balanced accuracy} & \multicolumn{3}{c}{AUC}\\
                \cline{2-7} 
                & $A_{loss}$ & $A_{black}$ & $\textsc{Parsing}$ & $A_{loss}$ & $A_{black}$ & $\textsc{Parsing}$\\
                \midrule
                PubMed\_RCT & 0.673$\pm$0.004 & 0.620$\pm$0.004 & \textbf{0.701$\pm$0.003} & 0.690$\pm$0.008 & 0.705$\pm$0.011 & \textbf{0.727$\pm$0.014}\\
                Yelp Reviews & 0.645$\pm$0.007 & 0.581$\pm$0.012 & \textbf{0.674$\pm$0.018} & 0.657$\pm$0.006 & 0.687$\pm$0.018 & \textbf{0.718$\pm$0.015}\\
                BC5CDR  & 0.660$\pm$0.011 & 0.618$\pm$0.005 & \textbf{0.692$\pm$0.003} & 0.693$\pm$0.015 & 0.720$\pm$0.005 & \textbf{0.734$\pm$0.003}\\
                PubMedQA & / & / & \textbf{/} & / & / & \textbf{/}\\
                Wiki Toxicity & 0.662$\pm$0.005 & 0.612$\pm$0.014 & \textbf{0.679$\pm$0.005} & 0.685$\pm$0.011 & 0.695$\pm$0.012 & \textbf{0.716$\pm$0.006}\\
                \cline{1-7} 
                \rule{0pt}{10pt}AG News & 0.645$\pm$0.004 & 0.596$\pm$0.012 & \textbf{0.652$\pm$0.002} & 0.656$\pm$0.004 & 0.663$\pm$0.012 & \textbf{0.700$\pm$0.008}\\
                Sentiment140 & 0.623$\pm$0.005 & 0.573$\pm$0.006 & \textbf{0.653$\pm$0.013} & 0.633$\pm$0.005 & 0.669$\pm$0.009 & \textbf{0.674$\pm$0.012}\\
                CoNLL-2003 & 0.659$\pm$0.003 & 0.622$\pm$0.014 & \textbf{0.674$\pm$0.002} & 0.686$\pm$0.010 & 0.706$\pm$0.013 & \textbf{0.724$\pm$0.014}\\
                \bottomrule
            \end{tabular}
            \caption{The maximum values achieved by \textsc{Parsing} and baselines in terms of \textit{balanced accuracy} and \textit{auc} when fine-tuning the GPT-2 medium model on different text tasks.}
        \end{table*}
        \begin{table*}
            \centering
            \setlength{\tabcolsep}{2mm}
            \fontsize{10pt}{14pt}\selectfont
            \begin{tabular}{l|ccc|ccc} 
                \toprule
                \multirow{2}{*}{Datasets}& \multicolumn{3}{c|}{Balanced accuracy} & \multicolumn{3}{c}{AUC}\\
                \cline{2-7} 
                & $A_{loss}$ & $A_{black}$ & $\textsc{Parsing}$ & $A_{loss}$ & $A_{black}$ & $\textsc{Parsing}$\\
                \midrule
                PubMed\_RCT & 0.693$\pm$0.002 & 0.632$\pm$0.007 & \textbf{0.717$\pm$0.004} & 0.706$\pm$0.002 & 0.724$\pm$0.010 & \textbf{0.740$\pm$0.013}\\
                Yelp Reviews & 0.683$\pm$0.004 & 0.619$\pm$0.013 & \textbf{0.703$\pm$0.008} & 0.699$\pm$0.009 & 0.724$\pm$0.016 & \textbf{0.728$\pm$0.010}\\
                BC5CDR  & 0.709$\pm$0.006 & 0.639$\pm$0.012 & \textbf{0.719$\pm$0.008} & 0.720$\pm$0.009 & 0.721$\pm$0.010 & \textbf{0.754$\pm$0.014}\\
                PubMedQA & 0.698$\pm$0.003 & 0.662$\pm$0.008 & \textbf{0.737$\pm$0.011} & 0.720$\pm$0.008 & 0.746$\pm$0.011 & \textbf{0.769$\pm$0.012}\\
                Wiki Toxicity & 0.681$\pm$0.004 & 0.637$\pm$0.014 & \textbf{0.699$\pm$0.017} & 0.711$\pm$0.008 & 0.732$\pm$0.015 & \textbf{0.741$\pm$0.010}\\
                \cline{1-7} 
                \rule{0pt}{10pt}AG News & 0.658$\pm$0.002 & 0.614$\pm$0.004 & \textbf{0.683$\pm$0.013} & 0.684$\pm$0.004 & 0.692$\pm$0.009 & \textbf{0.715$\pm$0.014}\\
                Sentiment140 & 0.654$\pm$0.006 & 0.607$\pm$0.004 & \textbf{0.663$\pm$0.006} & 0.653$\pm$0.012 & 0.638$\pm$0.008 & \textbf{0.685$\pm$0.009}\\
                CoNLL-2003 & 0.691$\pm$0.002 & 0.654$\pm$0.012 & \textbf{0.702$\pm$0.005} & 0.706$\pm$0.004 & 0.730$\pm$0.009 & \textbf{0.753$\pm$0.011}\\
                \bottomrule
            \end{tabular}
            \caption{The maximum values achieved by \textsc{Parsing} and baselines in terms of \textit{balanced accuracy} and \textit{auc} when fine-tuning the GPT-2 large model on different text tasks.}
        \end{table*}
        \begin{table*}
            \centering
            \setlength{\tabcolsep}{2mm}
            \fontsize{10pt}{14pt}\selectfont
            \begin{tabular}{l|ccc|ccc} 
                \toprule
                \multirow{2}{*}{Datasets}& \multicolumn{3}{c|}{Balanced accuracy} & \multicolumn{3}{c}{AUC}\\
                \cline{2-7} 
                & $A_{loss}$ & $A_{black}$ & $\textsc{Parsing}$ & $A_{loss}$ & $A_{black}$ & $\textsc{Parsing}$\\
                \midrule
                PubMed\_RCT & 0.716$\pm$0.004 & 0.647$\pm$0.014 & \textbf{0.734$\pm$0.007} & 0.746$\pm$0.008 & 0.746$\pm$0.015 & \textbf{0.758$\pm$0.012}\\
                Yelp Reviews & 0.694$\pm$0.007 & 0.611$\pm$0.011 & \textbf{0.704$\pm$0.004} & 0.716$\pm$0.010 & 0.701$\pm$0.013 & \textbf{0.734$\pm$0.009}\\
                BC5CDR  & 0.723$\pm$0.003 & 0.652$\pm$0.015 & \textbf{0.733$\pm$0.005} & 0.750$\pm$0.004 & 0.722$\pm$0.017 & \textbf{0.777$\pm$0.012}\\
                PubMedQA & 0.709$\pm$0.003 & 0.683$\pm$0.008 & \textbf{0.762$\pm$0.011} & 0.742$\pm$0.010 & 0.750$\pm$0.011 & \textbf{0.769$\pm$0.013}\\
                Wiki Toxicity & 0.703$\pm$0.004 & 0.653$\pm$0.008 & \textbf{0.713$\pm$0.008} & 0.737$\pm$0.009 & 0.718$\pm$0.014 & \textbf{0.755$\pm$0.015}\\
                \bottomrule
            \end{tabular}
            \caption{The maximum values achieved by \textsc{Parsing} and baselines in terms of \textit{balanced accuracy} and \textit{auc} when fine-tuning the GPT-Neo-1.3b model on different text tasks.}
        \end{table*}
        \begin{table*}
            \centering
            \setlength{\tabcolsep}{2mm}
            \fontsize{10pt}{14pt}\selectfont
            \begin{tabular}{l|ccc|ccc} 
                \toprule
                \multirow{2}{*}{Datasets}& \multicolumn{3}{c|}{Balanced accuracy} & \multicolumn{3}{c}{AUC}\\
                \cline{2-7} 
                & $A_{loss}$ & $A_{black}$ & $\textsc{Parsing}$ & $A_{loss}$ & $A_{black}$ & $\textsc{Parsing}$\\
                \midrule
                PubMed\_RCT & 0.732$\pm$0.002 & 0.684$\pm$0.009 & \textbf{0.751$\pm$0.007} & 0.771$\pm$0.007 & 0.757$\pm$0.014 & \textbf{0.794$\pm$0.017}\\
                Yelp Reviews & 0.711$\pm$0.004 & 0.637$\pm$0.016 & \textbf{0.719$\pm$0.006} & 0.730$\pm$0.009 & 0.705$\pm$0.011 & \textbf{0.740$\pm$0.013}\\
                BC5CDR  & 0.744$\pm$0.004 & 0.683$\pm$0.013 & \textbf{0.762$\pm$0.009} & 0.769$\pm$0.007 & 0.769$\pm$0.016 & \textbf{0.792$\pm$0.012}\\
                PubMedQA & 0.732$\pm$0.007 & 0.703$\pm$0.006 & \textbf{0.767$\pm$0.006} & 0.745$\pm$0.010 & 0.770$\pm$0.011 & \textbf{0.794$\pm$0.012}\\
                Wiki Toxicity & 0.713$\pm$0.004 & 0.677$\pm$0.008 & \textbf{0.733$\pm$0.011} & 0.740$\pm$0.004 & 0.754$\pm$0.014 & \textbf{0.769$\pm$0.012}\\
                \bottomrule
            \end{tabular}
            \caption{The maximum values achieved by \textsc{Parsing} and baselines in terms of \textit{balanced accuracy} and \textit{auc} when fine-tuning the GPT-Neo-2.7b model on different text tasks.}
        \end{table*}
        \begin{table*}
            \centering
            \setlength{\tabcolsep}{2mm}
            \fontsize{10pt}{14pt}\selectfont
            \begin{tabular}{l|ccc|ccc} 
                \toprule
                \multirow{2}{*}{Datasets}& \multicolumn{3}{c|}{Balanced accuracy} & \multicolumn{3}{c}{AUC}\\
                \cline{2-7} 
                & $A_{loss}$ & $A_{black}$ & $\textsc{Parsing}$ & $A_{loss}$ & $A_{black}$ & $\textsc{Parsing}$\\
                \midrule
                PubMed\_RCT & 0.748$\pm$0.004 & 0.694$\pm$0.012 & \textbf{0.766$\pm$0.009} & 0.782$\pm$0.008 & 0.778$\pm$0.012 & \textbf{0.794$\pm$0.012}\\
                Yelp Reviews & 0.727$\pm$0.007 & 0.643$\pm$0.010 & \textbf{0.733$\pm$0.013} & 0.750$\pm$0.011 & 0.736$\pm$0.013 & \textbf{0.779$\pm$0.012}\\
                BC5CDR  & / & / & \textbf{/} & / & / & \textbf{/}\\
                PubMedQA & 0.763$\pm$0.009 & 0.691$\pm$0.013 & \textbf{0.773$\pm$0.007} & 0.760$\pm$0.004 & 0.755$\pm$0.015 & \textbf{0.794$\pm$0.011}\\
                Wiki Toxicity & 0.731$\pm$0.006 & 0.683$\pm$0.015 & \textbf{0.733$\pm$0.014} & 0.753$\pm$0.010 & 0.758$\pm$0.012 & \textbf{0.777$\pm$0.012}\\
                \bottomrule
            \end{tabular}
            \caption{The maximum values achieved by \textsc{Parsing} and baselines in terms of \textit{balanced accuracy} and \textit{auc} when fine-tuning the Llama2-7b model on different text tasks.}
        \end{table*}
        \begin{table}
            \centering
            \setlength{\tabcolsep}{5mm}
            \fontsize{10pt}{14pt}\selectfont
            \begin{tabular}{l|ccc} 
                \toprule
                \multirow{2}{*}{Datasets}& \multicolumn{3}{c}{TPR at 0.1 FPR} \\
                \cline{2-4} 
                & $A_{loss}$ & $A_{black}$ & $\textsc{Parsing}$ \\
                \midrule
                PubMed\_RCT & 0.237$\pm$0.047 & 0.291$\pm$0.062 & \textbf{0.335$\pm$0.087}\\
                Yelp Reviews & 0.217$\pm$0.043 & 0.263$\pm$0.077 & \textbf{0.294$\pm$0.053}\\
                BC5CDR  & 0.233$\pm$0.067 & 0.293$\pm$0.083 & \textbf{0.333$\pm$0.091}\\
                PubMedQA & / & / & \textbf{/}\\
                Wiki Toxicity & 0.195$\pm$0.033 & 0.273$\pm$0.079 & \textbf{0.303$\pm$0.083} \\
                \cline{1-4} 
                \rule{0pt}{10pt}AG News & 0.183$\pm$0.024 & 0.255$\pm$0.077 & \textbf{0.280$\pm$0.079}\\
                Sentiment140 & 0.173$\pm$0.055 & 0.213$\pm$0.088  &  \textbf{0.241$\pm$0.093}\\
                CoNLL-2003 & 0.223$\pm$0.027 & 0.284$\pm$0.083 & \textbf{0.304$\pm$0.083}\\
                \bottomrule
            \end{tabular}
            \caption{The maximum values achieved by \textsc{Parsing} and baselines in terms of \textit{TPR at 0.1 FPR} when fine-tuning the GPT-2 medium model on different text tasks.}
        \end{table}
        \begin{table}
            \centering
            \setlength{\tabcolsep}{5mm}
            \fontsize{10pt}{12pt}\selectfont
            \begin{tabular}{l|ccc} 
                \toprule
                \multirow{2}{*}{Datasets}& \multicolumn{3}{c}{TPR at 0.1 FPR} \\
                \cline{2-4} 
                & $A_{loss}$ & $A_{black}$ & $\textsc{Parsing}$ \\
                \midrule
                PubMed\_RCT & 0.251$\pm$0.043 & 0.324$\pm$0.039 & \textbf{0.355$\pm$0.024}\\
                Yelp Reviews & 0.226$\pm$0.019 & 0.283$\pm$0.067 & \textbf{0.317$\pm$0.072}\\
                BC5CDR  & 0.253$\pm$0.031 & 0.326$\pm$0.055 & \textbf{0.367$\pm$0.055}\\
                PubMedQA & 0.276$\pm$0.044 & 0.343$\pm$0.083 & \textbf{0.387$\pm$0.063}\\
                Wiki Toxicity & 0.237$\pm$0.023 & 0.279$\pm$0.043 & \textbf{0.333$\pm$0.052} \\
                \cline{1-4} 
                AG News & 0.208$\pm$0.033 & 0.272$\pm$0.041  &  \textbf{0.285$\pm$0.017}\\
                Sentiment140 & 0.194$\pm$0.033 & 0.225$\pm$0.037 & \textbf{0.261$\pm$0.073}\\
                CoNLL-2003 & 0.237$\pm$0.041 & 0.307$\pm$0.067 & \textbf{0.324$\pm$0.080}\\
                \bottomrule
            \end{tabular}
            \caption{The maximum values achieved by \textsc{Parsing} and baselines in terms of \textit{TPR at 0.1 FPR} when fine-tuning the GPT-2 large model on different text tasks.}
        \end{table}
        \begin{table}
            \centering
            \setlength{\tabcolsep}{5mm}
            \fontsize{10pt}{12pt}\selectfont
            \begin{tabular}{l|ccc} 
                \toprule
                \multirow{2}{*}{Datasets}& \multicolumn{3}{c}{TPR at 0.1 FPR} \\
                \cline{2-4} 
                & $A_{loss}$ & $A_{black}$ & $\textsc{Parsing}$ \\
                \midrule
                PubMed\_RCT & 0.251$\pm$0.024 & 0.324$\pm$0.083 & \textbf{0.373$\pm$0.067}\\
                Yelp Reviews & 0.193$\pm$0.012 & 0.307$\pm$0.052 & \textbf{0.320$\pm$0.073}\\
                BC5CDR & 0.231$\pm$0.038 & 0.312$\pm$0.078 & \textbf{0.368$\pm$0.070}\\
                PubMedQA & 0.267$\pm$0.033 & 0.373$\pm$0.062 & \textbf{0.387$\pm$0.088}\\
                Wiki Toxicity & 0.193$\pm$0.021 & 0.320$\pm$0.063 & \textbf{0.333$\pm$0.057} \\
                \bottomrule
            \end{tabular}
            \caption{The maximum values achieved by \textsc{Parsing} and baselines in terms of \textit{TPR at 0.1 FPR} when fine-tuning the GPT-Neo-1.3b model on different text tasks.}
        \end{table}
        \begin{table}
            \centering
            \setlength{\tabcolsep}{5mm}
            \fontsize{10pt}{12pt}\selectfont
            \begin{tabular}{l|ccc} 
                \toprule
                \multirow{2}{*}{Datasets}& \multicolumn{3}{c}{TPR at 0.1 FPR} \\
                \cline{2-4} 
                & $A_{loss}$ & $A_{black}$ & $\textsc{Parsing}$ \\
                \midrule
                PubMed\_RCT & 0.263$\pm$0.035 & 0.367$\pm$0.091 & \textbf{0.397$\pm$0.085}\\
                Yelp Reviews & 0.213$\pm$0.029 & 0.324$\pm$0.061 & \textbf{0.351$\pm$0.011}\\
                BC5CDR  & 0.267$\pm$0.043 & 0.371$\pm$0.058 & \textbf{0.401$\pm$0.049}\\
                PubMedQA & 0.278$\pm$0.016 & 0.383$\pm$0.093 & \textbf{0.417$\pm$0.081}\\
                Wiki Toxicity & 0.251$\pm$0.033 & 0.327$\pm$0.056 & \textbf{0.353$\pm$0.079} \\
                \bottomrule
            \end{tabular}
            \caption{The maximum values achieved by \textsc{Parsing} and baselines in terms of \textit{TPR at 0.1 FPR} when fine-tuning the GPT-Neo-2.7b model on different text tasks.}
        \end{table}
        \begin{table}
            \centering
            \setlength{\tabcolsep}{5mm}
            \fontsize{10pt}{12pt}\selectfont
            \begin{tabular}{l|ccc} 
                \toprule
                \multirow{2}{*}{Datasets}& \multicolumn{3}{c}{TPR at 0.1 FPR} \\
                \cline{2-4} 
                & $A_{loss}$ & $A_{black}$ & $\textsc{Parsing}$ \\
                \midrule
                PubMed\_RCT & 0.263$\pm$0.043 & 0.387$\pm$0.063 & \textbf{0.403$\pm$0.067}\\
                Yelp Reviews & 0.223$\pm$0.009 & 0.357$\pm$0.047 & \textbf{0.371$\pm$0.044}\\
                BC5CDR  & / & / & \textbf{/}\\
                PubMedQA & 0.261$\pm$0.032 & 0.381$\pm$0.075 & \textbf{0.413$\pm$0.081 }\\
                Wiki Toxicity & 0.248$\pm$0.055 & 0.330$\pm$0.076  & \textbf{0.379$\pm$0.080} \\
                \bottomrule
            \end{tabular}
            \caption{The maximum values achieved by \textsc{Parsing} and baselines in terms of \textit{TPR at 0.1 FPR} when fine-tuning the Llama2-7b model on different text tasks.}
        \end{table}

        \subsection{Audits during Fine-tuning}
        Figures 8 to 11 present the audit results on the \textit{balance accuracy} metric during the 40 epochs of fine-tuning various sizes of models on the PubMed\_RCT, Yelp Reviews, BC5CDR, and PubMedQA, respectively. Through the visualized audit process, the changes in privacy risks encountered during the fine-tuning of language models are clearly visible. Figure 12 presents the audit results of the fine-tuning of four tasks in Llama2-7b. In the early stages of fine-tuning, QA tasks show lower privacy leakage risks, yet as training continues, the peak values of the quantified metrics for QA tasks exceed those of relatively simpler tasks. Moreover, compared to other smaller models, the peak for Llama2-7b appears very early and stabilizes after a slight decrease.
        \begin{figure}
            \centering
            \begin{minipage}{0.47\textwidth}
                \centering
                \includegraphics[width=\textwidth]{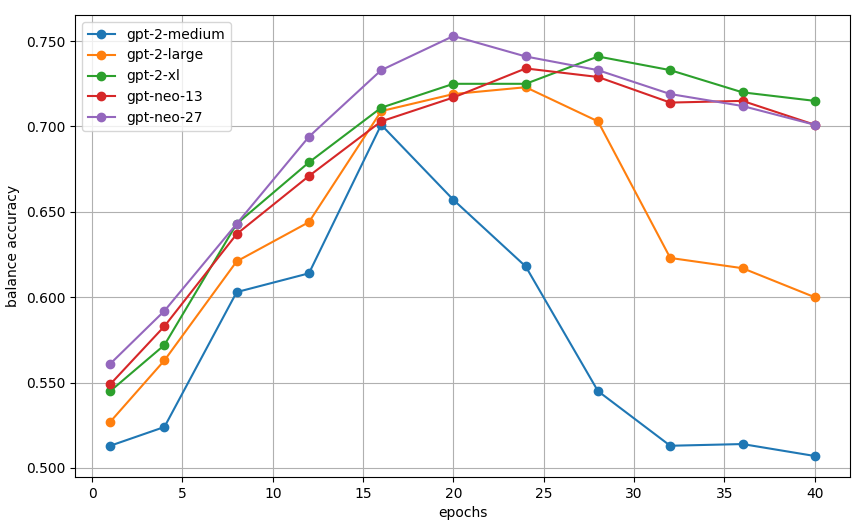}
                \caption{Balance accuracy audit results during the 40 epochs of fine-tuning each model on the PubMed\_RCT dataset.}
            \end{minipage}
            \hspace{0.01\textwidth}
            \begin{minipage}{0.47\textwidth}
                \centering
                \includegraphics[width=\textwidth]{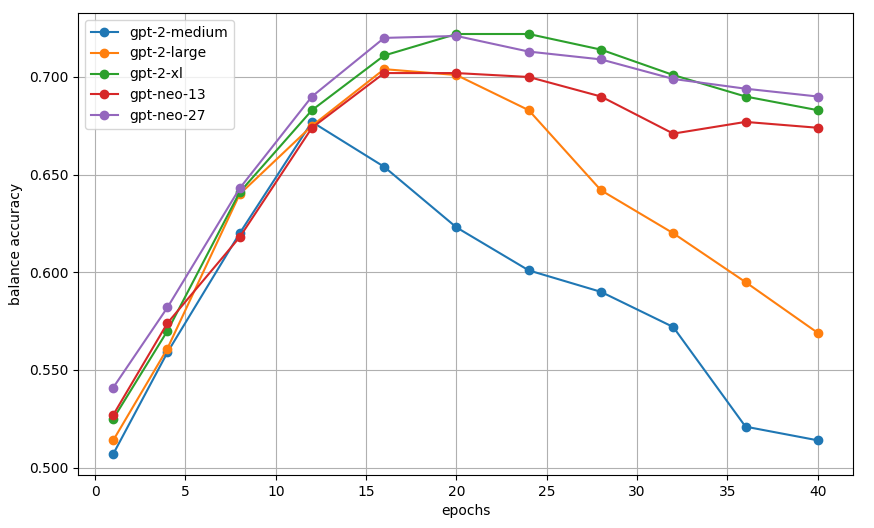}
                \caption{Balance accuracy audit results during the 40 epochs of fine-tuning each model on the Yelp Reviews dataset.}
            \end{minipage}
        \end{figure}

        \begin{figure}
            \centering
            \begin{minipage}{0.47\textwidth}
                \centering
                \includegraphics[width=\textwidth]{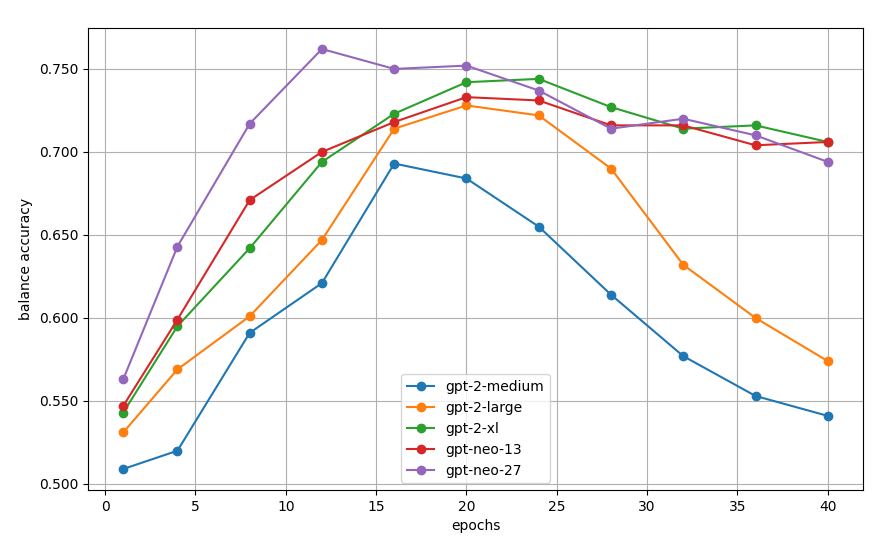}
                \caption{Balance accuracy audit results during the 40 epochs of fine-tuning each model on the BC5CDR dataset.}
            \end{minipage}
            \hspace{0.01\textwidth}
            \begin{minipage}{0.47\textwidth}
                \centering
                \includegraphics[width=\textwidth]{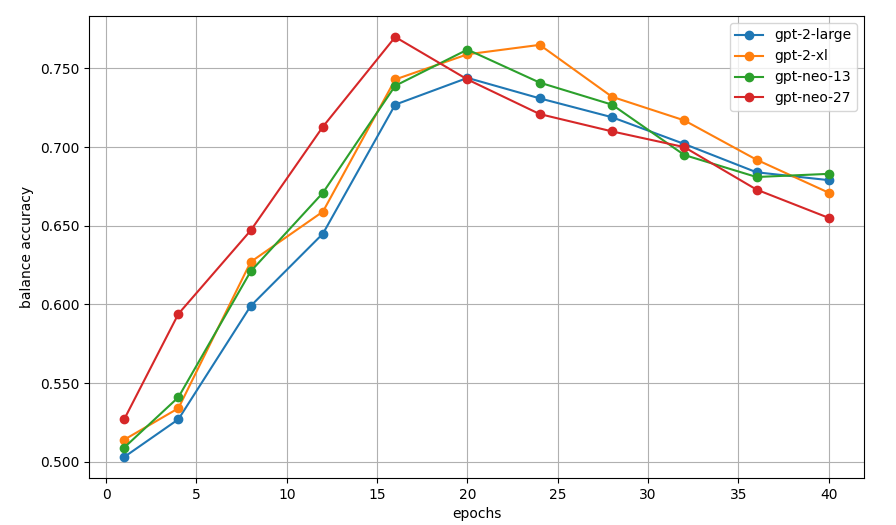}
                \caption{Balance accuracy audit results during the 40 epochs of fine-tuning each model on the PubMedQA dataset.}
            \end{minipage}
        \end{figure}
        
        \begin{figure}
          \centering
          \includegraphics[width=0.7\linewidth]{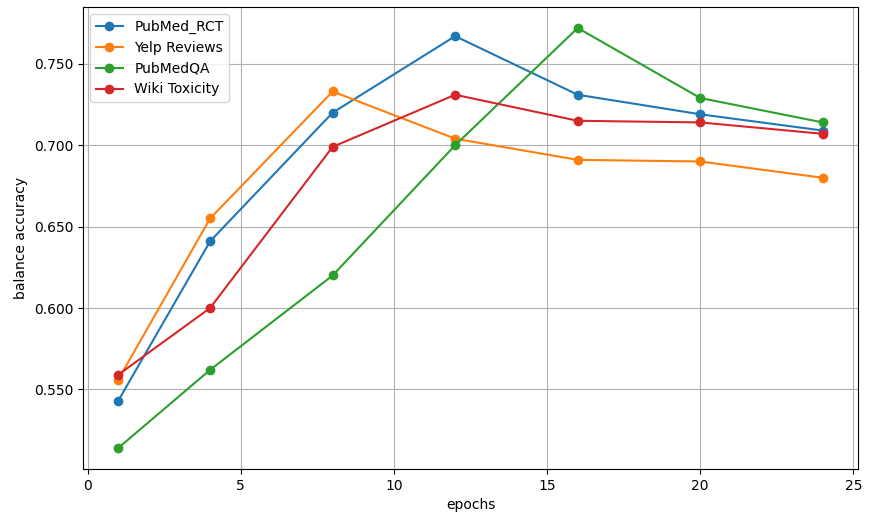}
          \caption{Balance accuracy audit results during the 40 epochs of fine-tuning Llama2-7b on various datasets.}
        \end{figure}
        
        \subsection{Visualizing Privacy Risks}
        \begin{figure}[t]
          \centering
          \includegraphics[width=0.6\linewidth]{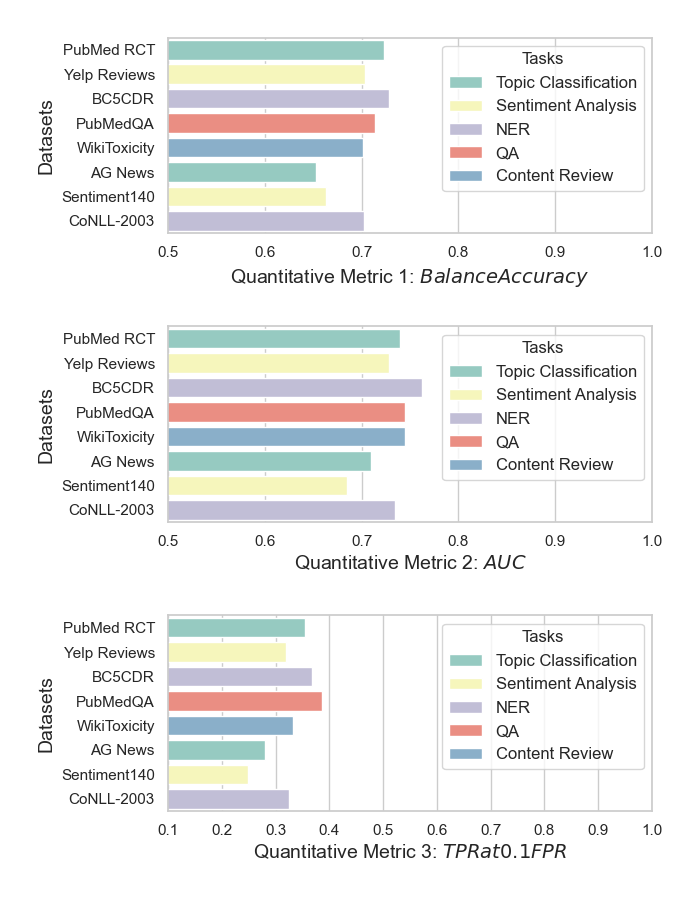}
          \caption{Visualizing the quantified privacy risk results.}
        \end{figure}
        To further uncover privacy risks during the model fine-tuning process, the quantified risk results can be visualized as shown in Figure 13. The starting points for $accuracy$ and $auc$ are 0.5, and for $TPR_{0.1}$ it is 0.1. This visualization can be extended to each epoch during fine-tuning, providing a clearer view of the upper bounds of privacy leakage risks throughout the language model fine-tuning process.

        \section{Further Discussion}
        \label{sec:appendix}
        
        \subsection{Factors Affecting MIA Vulnerability}
        Apart from specific text tasks, text lengths, model sizes, and training iterations, during the experiments we also discovered that the batch size set for fine-tuning training impacts the values of the privacy leakage audit metrics.
        \begin{itemize}
            \item Text tasks - Higher complexity tasks show greater privacy leakage risks after overtraining.
       
            \item Text length - Longer texts may pose greater privacy leakage risks after overtraining.
        
            \item Model size - Larger models are prone to leaking more information.
        
            \item Iterations - In the early stages of training, privacy leakage risk is low, but as training continues, the privacy leakage level peaks and then gradually declines to a convergence point.	
        
            \item Trainable parameters — The fewer parameters fine-tuned during training, the lower the risk of privacy leakage under normal circumstances.	
        
            \item Batch size - Larger fine-tuned batch sizes during training, result in lower privacy leakage risks.	
        \end{itemize}

        \subsection{Practical Privacy Mitigation Techniques}
        Additionally, we provide alternatives for privacy mitigation that potentially leads to future exploration in the direction of active privacy auditing.
        \begin{itemize}
            \item Differential Privacy (DP) - DP is a strong theoretical framework that adds noise to the training process to protect individual data points. However, DP often leads to a trade-off between privacy and model accuracy.
       
            \item Model distillation - Distillation involves training a smaller ``student" model to replicate the behavior of a larger ``teacher" model. This can reduce the risk of overfitting and memorization in the student model.
        
            \item Data augmentation - Data augmentation and adversarial training are also effective in improving model generalization.	
        
            \item Batch size - Larger batch sizes result in lower privacy leakage risks.	
        \end{itemize}

\end{document}